\pdfoutput=1
\PassOptionsToPackage{prologue,dvipsnames}{xcolor}

\documentclass[11pt]{article}
\usepackage{authblk}
\usepackage{EMNLP2023}

\usepackage{times}
\usepackage{latexsym}
\usepackage{multirow}
\usepackage{ctable}
\usepackage{adjustbox}
\usepackage{comment}
\usepackage{tcolorbox}
\usepackage[dvipsnames]{xcolor}
\usepackage{array}
\usepackage{xcolor}
\usepackage{pifont}
\usepackage{enumitem}
\usepackage[T1]{fontenc}

\usepackage[utf8]{inputenc}
\usepackage{microtype}

\usepackage{inconsolata}

\title{mRAG: Elucidating the Design Space of Multi-modal\\ Retrieval-Augmented Generation}

\author[1]{Chan-Wei Hu}
\author[2]{Yueqi Wang}
\author[1]{Shuo Xing}
\author[3]{Chia-Ju Chen}
\author[4]{Suofei Feng}
\author[5]{\\Ryan Rossi}
\author[1]{Zhengzhong Tu$^*$}
\affil[1]{Texas A\&M University}
\affil[2]{University of California, Berkeley}
\affil[3]{University of Texas at Austin}
\affil[4]{Stanford University}
\affil[5]{Adobe Research}
\title{mRAG: Elucidating the Design Space of Multi-modal\\ Retrieval-Augmented Generation}
\date{}

\begin{document}
\maketitle

\renewcommand{\thefootnote}{\fnsymbol{footnote}}
\footnotetext[1]{{Corresponding Author: Zhengzhong Tu (tzz@tamu.edu)}}

\begin{abstract}
Large Vision-Language Models (LVLMs) have made remarkable strides in multimodal tasks such as visual question answering, visual grounding, and complex reasoning.
However, they remain limited by static training data, susceptibility to hallucinations, and inability to verify claims against up-to-date, external evidence, compromising their performance in dynamic real-world applications.
Retrieval-Augmented Generation (RAG) offers a practical solution to mitigate these challenges by allowing the LVLMs to access large-scale knowledge databases via retrieval mechanisms, thereby grounding model outputs in factual, contextually relevant information.
Here in this paper, we conduct the first systematic dissection of the multimodal RAG pipeline for LVLMs, explicitly investigating (1) the retrieval phase: on the modality configurations and retrieval strategies, (2) the re-ranking stage: on strategies to mitigate positional biases and improve the relevance of retrieved evidence, and (3) the generation phase: we further investigate how to best integrate retrieved candidates into the final generation process.
%
%
Finally, we extend to explore a unified agentic framework that integrates re-ranking and generation through self-reflection, enabling LVLMs to select relevant evidence and suppress irrelevant context dynamically.
Our full-stack exploration of RAG for LVLMs yields substantial insights, resulting in an average performance boost of 5\% without any fine-tuning.
\end{abstract}
\begin{figure*}
  \centering\includegraphics[width=0.90\textwidth]{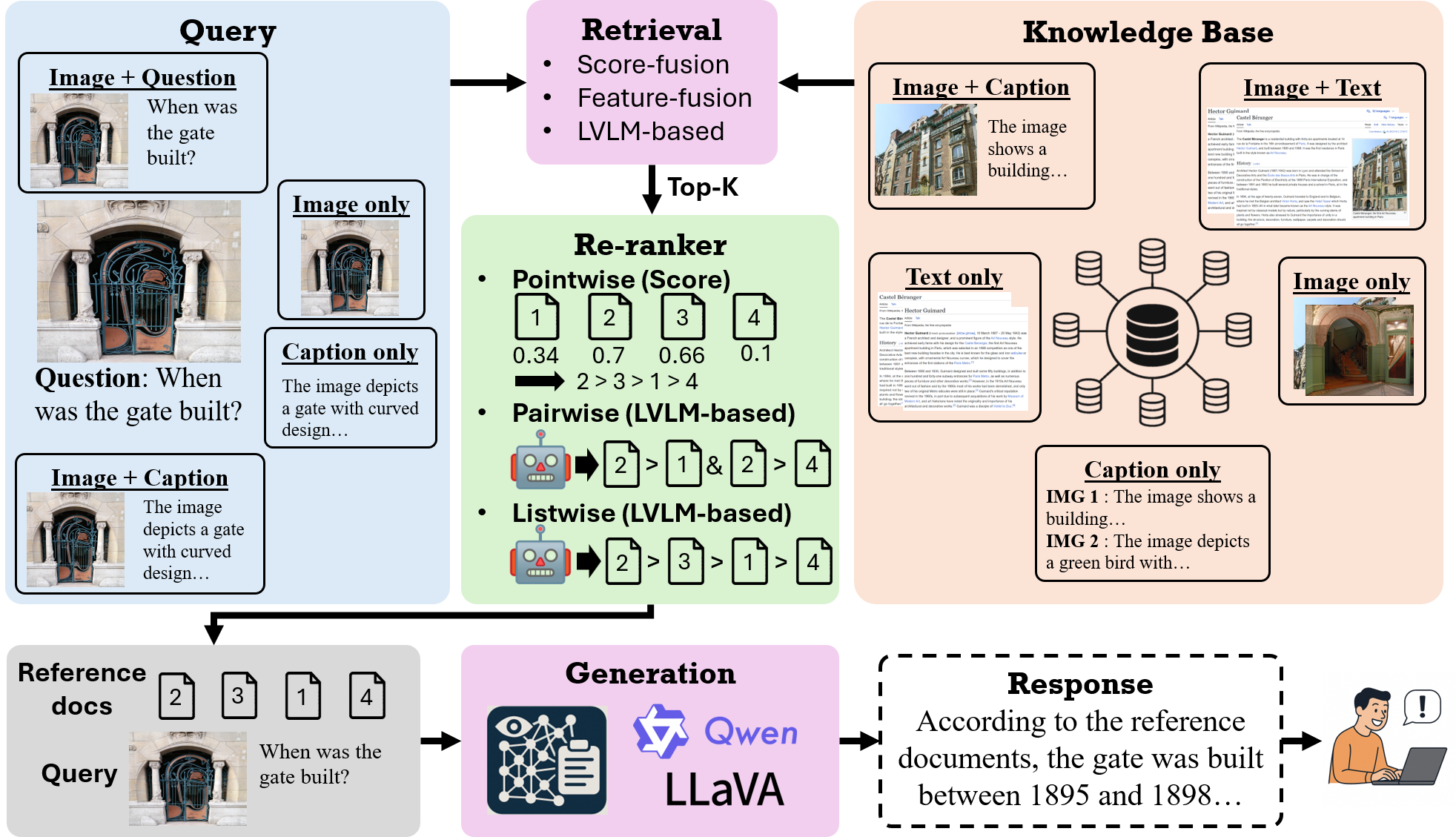}
  \vspace{-1mm}
  \caption{The multi-modal RAG (mRAG) pipeline utilized in our journey to exploit the design space of each component thereof: \ding{182} \textbf{Retrieval} (\S \ref{ret_section}), \ding{183} \textbf{Re-ranking} (\S \ref{sec:reranking}), and \ding{184} \textbf{Generation} (\S \ref{gen_section}).}
  \label{teaser}
  \vspace{-4mm}
\end{figure*}

\section{Introduction}
Recent advancements in Large Vision-Language Models (LVLMs) have significantly enhanced their capabilities in processing and generating multi-modal content, substantially benefiting real-world applications such as visual question answering (VQA)~\cite{vlm4vision, guiding, Qwen-VL, internvl}, visual grounding~\cite{vlm_grounding_1, vlm_grounding_2, vlm_grounding_3}, complex task planning~\cite{vlm_task_1, vlm_task_2, vlm_task_3}, and physical reasoning \cite{spatialvlm, phys_vlm_1, phys_vlm_2, phys_vlm_3}. 
Despite these remarkable strides, however, LVLMs inherently suffer from several fundamental limitations, primarily stemming from their reliance on static, frozen training data~\cite{abootorabi2025ask, mei2025survey}, insufficient semantic grounding capabilities~\cite{vlm_semantic_1, vlm_semantic_2}, and inadequate alignment across modalities~\cite{vlm_align_1, vlm_align_2}.
Specifically, these limitations lead to practical challenges, such as prone to producing factual hallucinations—outputs that appear plausible but are factually incorrect~\cite{vlm_hallucination_1, vlm_hallucination_2, vlm_hallucination_3, vlm_hallucination_4}, struggling with outdated knowledge for time-sensitive question answering~\cite{tsqa_1, tsqa_2, tsqa_3}, and lacking robust verification mechanisms to validate claims using external evidence~\cite{vlm_verify_1, vlm_verify_2, vlm_verify_3}. 
%

Retrieval-Augmented Generation (RAG) offers a promising and practical solution to mitigate these limitations by equipping LVLMs to access, retrieve, and integrate external, up-to-date knowledge sources~\cite{rag1, rag2, rag3}. 
Specifically, RAG incorporates retrieval mechanisms to fetch contextually relevant information from large-scale knowledge bases, significantly reducing the likelihood of factual hallucinations and enhancing the accuracy of generated outputs.
Recently, multi-modal extensions of RAG (referred to as \textbf{mRAG}) have emerged, integrating textual, visual, and other modalities into the retrieval-generation pipeline, substantially expanding the versatility of LVLMs across many domains. 
For instance, multimodal RAG has successfully enabled evidence-based medical diagnostics \cite{mmed, rule}, decision-making in autonomous driving \cite{rag-driver}, and industry applications~\cite {riedler2024beyond}.

\textbf{Prior Work.} Despite this rapid progress, existing research in multimodal RAG remains fragmented and lacks a comprehensive exploration of its full design space.
\underline{Firstly}, there exists limited empirical validation of how multi-modal alignment strategies impact retrieval effectiveness in mRAG workflows. 
While some studies propose fusion techniques combining visual and textual embeddings \cite{uniIR, ovis, MagicLens}, they do not analyze how these methods perform across different combinations of modalities.
\underline{Second}, re-ranking approaches in existing mRAG frameworks predominantly rely on straightforward relevance scoring mechanisms, assigning absolute scores based on query-candidate similarity~\cite{rerank_score}. Alternative ranking strategies, such as pairwise and listwise methods~\cite{first, pairwise, listwise, rerank_methods}, have remained underexplored in multimodal contexts.
\underline{Lastly}, current mRAG frameworks typically isolate the retrieval, re-ranking, and generation phases, resulting in suboptimal coordination between evidence selection and answer generation.

\textbf{Our Work.} To this end, we present a systematic study that elucidates the comprehensive design space of mRAG for LVLMs, methodically dissecting each critical phase of the mRAG pipeline.
We start with a baseline design as shown in Fig.~\ref{teaser}, and perform detailed investigations into: \ding{182} the \textbf{retrieval} phase, analyzing multiple modality configurations and retrieval strategies; 
\ding{183} the \textbf{re-ranking} phase, evaluating different approaches aimed at mitigating positional biases and enhancing evidence relevant; and \ding{184} the \textbf{generation} phase, exploring optimal methods for integrating retrieved candidates into the final model outputs.
Through this structured exploration, we identify crucial insights and best practices across each phase, ultimately converging on an \textbf{optimized mRAG pipeline} that involves the following recipe:
\vspace{-3pt}
\begin{tcolorbox}
\textbf{\textcolor{blue}{Recipe:}} integration of \textbf{(1)} EVA-CLIP as retriever, \textbf{(2)} listwise LVLM-based re-ranking, and \textbf{(3)} only providing most relevant document for generation yields a +2.32\%/+0.65\% response accuracy increase on benchmark datasets including E-VQA and InfoSeek.
\end{tcolorbox}

Finally, building on our best mRAG pipeline, we present an initial exploration into a unified \textbf{Agentic mRAG} framework by incorporating a self-reflection mechanism.
 We systematically compare multiple strategies, utilizing powerful LVLM-based re-rankers to enhance candidate ordering. Additionally, we examine how retrieval quality impacts answer accuracy, specifically highlighting how irrelevant candidates degrade performance even when correct information is present. Motivated by these insights, we propose a unified agentic framework that integrates re-ranking and generation via iterative self-reflection. This unified approach enables LVLMs to dynamically assess candidate relevance, selectively leveraging beneficial context while suppressing irrelevant information. 


\section{Preliminaries}
Before exploring best practices for mRAG, we provide an overview of the general dataset setup and the evaluation metrics.

\subsection{Dataset Constructions}
\textbf{Original Dataset.} Following prior studies \cite{wiki-llava, echosight}, we adopt VQA as the task of our study. We chose two knowledge-based VQA datasets.
\begin{itemize}
[leftmargin=*,noitemsep]
\vspace{-4pt}
  \item \textbf{Encyclopedic-VQA (E-VQA)} \cite{evqa} comprises 221k unique visual question-answer pairs. These images are sourced from iNaturalist 2021 \cite{inat} and Google Landmarks Dataset V2 \cite{gldv2}. The visual questions emphasize fine-grained category distinctions and instance-level recognition, requiring alignment between visual content and structured knowledge. A knowledge base of 2M Wikipedia articles is provided.
  \item \textbf{InfoSeek} \cite{infoseek} is designed to evaluate models on knowledge-intensive, information-seeking questions that cannot be answered using only visual content or common sense. It contains 1.3M curated image-question-answers corresponding to 100K Wikipedia articles. 
\end{itemize}

\begin{table}[!h]
\centering
\vspace{-3mm}
\caption{Statistics of the distilled dataset.} 
\vspace{-2mm}
\label{tab:date_stat}
\begin{adjustbox}{width=200pt}
\begin{tabular}{ccccc}
\specialrule{.2em}{.05em}{.05em}
\multirow{2}{*}{} & \multicolumn{2}{c}{InfoSeek} & \multicolumn{2}{c}{E-VQA} \\
\cline{2-5}
                  & Original     & Distilled     & Original    & Distilled   \\
\hline
\#articles        & 100K         & 50K           & 2M          & 50K         \\
\hline
\#images          & 371K         & 184K          & 6.6M        & 171K    \\ 
\specialrule{.2em}{.05em}{.05em}        
\end{tabular}
\vspace{-2mm}
\end{adjustbox}
\end{table}

\textbf{Distilled Dataset.} The scale of knowledge bases in E-VQA and InfoSeek introduces significant computational demands when employing LVLMs as retrievers, particularly for vector search operations, requiring full knowledge base encoding before retrieval phase. As a result, we distill a 50k-article subset through sampling. This process ensures that all evaluation queries remain answerable within the reduced knowledge base. Additionally, we sample articles in a manner that preserves the original category distribution. The statistics of the distilled datasets are presented in Table \ref{tab:date_stat}. For evaluation, in line with previous research \cite{echosight}, we use 4,750 test cases for E-VQA, and 5,000 cases for InfoSeek. More details are in Appendix \ref{app:dataset_details}.

\subsection{Evaluation Metrics}
\textbf{Retrieval.} We assess retrieval performance using Recall@$K$ and Mean Reciprocal Rank (MRR). Recall@$K$ measures the percentage of the correct article found in the top-$K$ retrieved candidates across the evaluation queries. MRR calculates the reciprocal of the rank at which the first correct article is retrieved. It provides a clear measure of how quickly a relevant article is found to ensure LVLMs receive critical contextual information early after the re-ranking phase. 
\\
\textbf{Visual Question Answering.} We evaluate VQA performance through complementary metrics addressing both lexical and semantic accuracy. ROUGE-L is used to compare the model response with reference answers. However, this metric may not fully capture answers that are phrased differently but convey the same meaning. To capture semantic correctness, we employ InternVL3 \cite{internvl, internvl3} and GPT-4.1 \cite{gpt4} as automated judges to assess if LVLM's answer is semantically correct, providing a more comprehensive evaluation of VQA performance.
All evaluations are performed using a temperature setting of 0.
\section{Retrieval Configurations and Strategies}
\label{ret_section}
The retrieval phase in mRAG requires careful consideration of input modalities and their alignment with candidates in the knowledge base because the complexity of processing heterogeneous data types, such as images and text, exhibits distinct semantic structures and embedding space distributions. This section systematically examines how different modality pairings between user queries and candidates in the knowledge base impact retrieval performance in mRAG. Unlike traditional approaches that rely on fine-tuned models adapted to specific knowledge bases, our investigation focuses on zero-shot retrieval capabilities using frozen pre-trained models. Furthermore, we evaluate multi-modal configurations where queries and candidates in the knowledge base may independently combine text, image, or hybrid modalities, testing models' inherent abilities to establish semantic alignment across modalities without parameter updates. 
Once the query and candidates' embeddings are generated, we employ the FAISS library \cite{faiss} with dot product similarity for retrieving
top-$K$ candidates and their corresponding wiki articles. 

\begin{table*}[!t]
\centering
\caption{Results on foundational stage. Following \cite{uniIR, mm-embed}, we report Recall@5 for both datasets. The $I \leftrightarrow IT$ modality configuration achieves peak Recall@5 scores (underlined) for both datasets using EVA-CLIP$_{SF}$, outperforming all other strategies including MLLM-based methods.} 
\vspace{-2mm}
\label{tab:fou_ret_result}
\begin{adjustbox}{width=400pt}
\begin{tabular}{c|c|llllll}
\specialrule{.2em}{.05em}{.05em} 
\multirow{2}{*}{\textbf{Dataset}}  & \multirow{2}{*}{\textbf{Task (Query $\leftrightarrow$ KB)}} & \multicolumn{6}{c}{\textbf{Retrieval Strategy}}   \\ \cline{3-8}  
                                   &  & \multicolumn{1}{c|}{CLIP$_{SF}$}    & \multicolumn{1}{c|}{EVA-CLIP$_{SF}$} & \multicolumn{1}{c|}{BGE-CLIP$_{SF}$} & \multicolumn{1}{c|}{BLIP$_{FF}$} & \multicolumn{1}{c|}{BGE-MLLM} & \multicolumn{1}{c}{GME} \\ 
\specialrule{.1em}{.05em}{.05em}
\multirow{4}{*}{InfoSeek} & $I \leftrightarrow I$ & \multicolumn{1}{c|}{67.5} & \multicolumn{1}{c|}{77.84} & \multicolumn{1}{c|}{49.92} & \multicolumn{1}{c|}{57.2} & \multicolumn{1}{c|}{39} & \multicolumn{1}{c}{56.52}      \\  
                          & $I \leftrightarrow IT$ & \multicolumn{1}{c|}{73.6}  & \multicolumn{1}{c|}{\underline{81.58}} & \multicolumn{1}{c|}{41.02} & \multicolumn{1}{c|}{64.22} & \multicolumn{1}{c|}{11.36} & \multicolumn{1}{c}{62.32}     \\ 
                          & $IQ \leftrightarrow I$ & \multicolumn{1}{c|}{67.5}  & \multicolumn{1}{c|}{77.8} & \multicolumn{1}{c|}{13.36} &  \multicolumn{1}{c|}{42.1} & \multicolumn{1}{c|}{18.78} & \multicolumn{1}{c}{74.94}      \\ 
                          & $IQ \leftrightarrow IT$ & \multicolumn{1}{c|}{27.2}  & \multicolumn{1}{c|}{76.94} & \multicolumn{1}{c|}{0.7} & \multicolumn{1}{c|}{33.92} & \multicolumn{1}{c|}{10.72} & \multicolumn{1}{c}{81.48}      \\ 
\hline
\multirow{4}{*}{E-VQA} & $I \leftrightarrow I$ & \multicolumn{1}{c|}{62.8} & \multicolumn{1}{c|}{75.9} & \multicolumn{1}{c|}{46.46} & \multicolumn{1}{c|}{54.3} & \multicolumn{1}{c|}{33.8} &                                  \multicolumn{1}{c}{53.6}      \\  
                          & $I \leftrightarrow IT$ & \multicolumn{1}{c|}{72.29}  & \multicolumn{1}{c|}{\underline{80.69}} & \multicolumn{1}{c|}{35.28} & \multicolumn{1}{c|}{59.81} & \multicolumn{1}{c|}{13.37} & \multicolumn{1}{c}{50.84}     \\ 
                          & $IQ \leftrightarrow I$ & \multicolumn{1}{c|}{63.3}  & \multicolumn{1}{c|}{76.75} & \multicolumn{1}{c|}{11.34} & \multicolumn{1}{c|}{40.44} & \multicolumn{1}{c|}{15.11} & \multicolumn{1}{c}{61.93}      \\ 
                          & $IQ \leftrightarrow IT$ & \multicolumn{1}{c|}{31.2}  & \multicolumn{1}{c|}{77.2} & \multicolumn{1}{c|}{6.8} & \multicolumn{1}{c|}{38.32} & \multicolumn{1}{c|}{21.35} & \multicolumn{1}{c}{77.03}      \\ 
\specialrule{.2em}{.05em}{.05em} 
\end{tabular}
\vspace{-2mm}
\end{adjustbox}
\end{table*}

\begin{table}[t]
\centering
\caption{Results on expansion stage using EVA-CLIP. Underlined scores are the best Recall@5 based on raw data, shown in Table \ref{tab:fou_ret_result}. } 
\vspace{-2mm}
\label{tab:exp_ret_result}
\begin{adjustbox}{width=220pt}
\begin{tabular}{c|c|lll}
\specialrule{.2em}{.05em}{.05em} 
\multirow{2}{*}{\textbf{Dataset}}  & \multirow{2}{*}{\textbf{Task (Query $\leftrightarrow$ KB)}} & \multicolumn{3}{c}{\textbf{Recall@K}}   \\ \cline{3-5}  
                                   &  & \multicolumn{1}{c|}{K=1}    & \multicolumn{1}{c|}{K=5} & \multicolumn{1}{c}{K=10}  \\ 
\specialrule{.1em}{.05em}{.05em}
\multirow{12}{*}{InfoSeek} & $I \leftrightarrow I$ & \multicolumn{1}{c|}{57.7} & \multicolumn{1}{c|}{77.84} & \multicolumn{1}{c}{82.08}   \\ 
                          & $I \leftrightarrow IT$ & \multicolumn{1}{c|}{\underline{63.42}} & \multicolumn{1}{c|}{\underline{81.58}} & \multicolumn{1}{c}{\underline{85.22}}    \\
                          & $I \leftrightarrow IC$ & \multicolumn{1}{c|}{56.32} & \multicolumn{1}{c|}{77.3} & \multicolumn{1}{c}{81.39}   \\  
                          & $I \leftrightarrow C$ & \multicolumn{1}{c|}{31.1}  & \multicolumn{1}{c|}{52.16} & \multicolumn{1}{c}{58.9}   \\ 
                          & $IC \leftrightarrow I$ & \multicolumn{1}{c|}{56.12} & \multicolumn{1}{c|}{76.63} & \multicolumn{1}{c}{81.21}   \\
                          & $IC \leftrightarrow IC$ & \multicolumn{1}{c|}{53.02} & \multicolumn{1}{c|}{74.1} & \multicolumn{1}{c}{79.23}   \\  
                          & $IC \leftrightarrow C$ & \multicolumn{1}{c|}{23.9}  & \multicolumn{1}{c|}{42.23} & \multicolumn{1}{c}{49.74}   \\
                          & $IC \leftrightarrow IT$ & \multicolumn{1}{c|}{\textbf{64.44}} & \multicolumn{1}{c|}{\textbf{82.43}} & \multicolumn{1}{c}{\textbf{85.32}}   \\
                          & $C \leftrightarrow I$ & \multicolumn{1}{c|}{21.69} & \multicolumn{1}{c|}{38.97} & \multicolumn{1}{c}{46.3}   \\
                          & $C \leftrightarrow IC$ & \multicolumn{1}{c|}{21.75} & \multicolumn{1}{c|}{38.91} & \multicolumn{1}{c}{46.56}    \\  
                          & $C \leftrightarrow C$ & \multicolumn{1}{c|}{17.77}  & \multicolumn{1}{c|}{32.27} & \multicolumn{1}{c}{39.01}  \\
                          & $C \leftrightarrow IT$ & \multicolumn{1}{c|}{26.71}  & \multicolumn{1}{c|}{42.4} & \multicolumn{1}{c}{48.78}  \\
\hline
\multirow{12}{*}{E-VQA}   & $I \leftrightarrow I$ & \multicolumn{1}{c|}{54.5} & \multicolumn{1}{c|}{75.9} & \multicolumn{1}{c}{81}   \\
                          & $I \leftrightarrow IT$ & \multicolumn{1}{c|}{\underline{61.85}} & \multicolumn{1}{c|}{\underline{80.69}} & \multicolumn{1}{c}{\underline{85.51}}    \\
                          & $I \leftrightarrow IC$ & \multicolumn{1}{c|}{54.32} & \multicolumn{1}{c|}{75.12} & \multicolumn{1}{c}{80.63}   \\  
                          & $I \leftrightarrow C$ & \multicolumn{1}{c|}{23.77}  & \multicolumn{1}{c|}{41.18} & \multicolumn{1}{c}{48.88}   \\ 
                          & $IC \leftrightarrow I$ & \multicolumn{1}{c|}{53.26} & \multicolumn{1}{c|}{75.71} & \multicolumn{1}{c}{80.36}   \\
                          & $IC \leftrightarrow IC$ & \multicolumn{1}{c|}{49.26} & \multicolumn{1}{c|}{70.99} & \multicolumn{1}{c}{78.04}   \\  
                          & $IC \leftrightarrow C$ & \multicolumn{1}{c|}{20.42}  & \multicolumn{1}{c|}{34.82} & \multicolumn{1}{c}{42.59}   \\
                          & $IC \leftrightarrow IT$ & \multicolumn{1}{c|}{\textbf{62.61}} & \multicolumn{1}{c|}{\textbf{80.8}} & \multicolumn{1}{c}{\textbf{86.47}}   \\
                          & $C \leftrightarrow I$ & \multicolumn{1}{c|}{17.81} & \multicolumn{1}{c|}{30.86} & \multicolumn{1}{c}{38.46}   \\
                          & $C \leftrightarrow IC$ & \multicolumn{1}{c|}{18.84} & \multicolumn{1}{c|}{34.8} & \multicolumn{1}{c}{41.62}    \\  
                          & $C \leftrightarrow C$ & \multicolumn{1}{c|}{12.61}  & \multicolumn{1}{c|}{25.89} & \multicolumn{1}{c}{33.05}  \\
                          & $C \leftrightarrow IT$ & \multicolumn{1}{c|}{19.2}  & \multicolumn{1}{c|}{33.81} & \multicolumn{1}{c}{42}  \\
\specialrule{.2em}{.05em}{.05em} 
\end{tabular}
\end{adjustbox}
\vspace{-4mm}
\end{table}

\subsection{Modality Configurations}
Retrieval effectiveness depends on how information is encoded in queries and candidates in the knowledge base. In this study, we investigate five modality configurations to systematically evaluate multimodal retrieval.
The \textbf{image-only ($I$)} configuration relies solely on images and is applicable to both query and candidate sides. The \textbf{image + question ($IQ$)} setting, which combines image with the user’s question to enable joint vision-language reasoning, is used solely on the query side, as questions are inherently tied to the query image and not present in the knowledge base. The \textbf{image + text ($IT$)} configuration fuses images with associated textual information, such as article passages, and is only applied to knowledge base candidates. The \textbf{image + caption ($IC$)} setup augments images with generated captions from a caption model, providing explicit semantic cues to complement the image; this configuration is applicable to both queries and candidates. Finally, the \textbf{caption-only ($C$)} configuration uses generated captions alone, and can also be applied to either side. We do not consider a question-only configuration, as questions are always grounded in the corresponding query image.

\subsection{Retrieval Strategies}
\textbf{Score Fusion}. This approach involves combining scores derived from different modalities. For instance, $CLIP_{SF}$ in \cite{uniIR} employs a dual encoder where visual and textual modalities are processed independently through separate unimodal encoders, producing two distinct embedding vectors of the same dimensionality. The fusion mechanism operates by computing a weighted linear combination of these unimodal embeddings to produce a unified representation vector. Formally, given a visual encoder $\phi_{vis}$ and a text encoder $\phi_{txt}$, the fusion score $S$ is shown as $S = \phi_{vis}(I) + \phi_{txt}(\tau)$, where $\tau \in \{C,T,Q\}$ depending on the configuration of the modality. 
\\
\textbf{Feature Fusion}. Unlike score fusion, which combines modality-specific similarities post-hoc, feature fusion integrates multimodal features during the encoding phase. This fusion approach generates a single feature representation for multi-modal queries or candidates by applying mixed-modality layers. However, this approach typically requires fine-tuning the fusion layers on the target knowledge base, while our work prioritizes evaluating frozen models' inherent ability to bridge modality gaps using their pretrained representations. Therefore, we utilize a pretrained $BLIP_{FF}$ model from \cite{uniIR}. $BLIP_{FF}$ is trained on diverse datasets and is capable of retrieving heterogeneous outputs in both text and image modalities.
\\
\textbf{LVLM-based Retriever}. 
Modern LVLMs integrate visual encoders (typically vision transformers) with pretrained language models, enabling them to natively process image-text token sequences. This ability makes them particularly suited for zero-shot retrieval scenarios where frozen pretrained parameters must bridge modality gaps and map diverse modalities into a unified token space. Our evaluation focuses on benchmarking these retrieval models' \cite{mm-embed, e5-v, bge, gme} zero-shot retrieval performance across modality configurations, testing their ability to align queries and candidates without task-specific fine-tuning.

In this study, we evaluated the retrieval performance of six distinct approaches across three types above, including three score fusion methods: CLIP$_{SF}$ \cite{uniIR}, EVA-CLIP$_{SF}$ \cite{evaclip-1, evaclip-2}, and BGE-CLIP$_{SF}$ \cite{bge} (For score fusion, image and text embeddings are assigned equal weight), one feature fusion BLIP$_{FF}$ \cite{uniIR}, and two LVLM-based retrievers, BGE-MLLM \cite{bge} and GME \cite{gme}.

\subsection{Results}
Our evaluation follows a two-stage design to isolate and quantify the impact of different modality combinations on retrieval performance. The \textbf{foundational stage} establishes baseline performance by evaluating three core configurations: $I$, $IT$, and $IQ$, all without caption augmentation. This stage identifies which retrieval strategies perform the best on raw inputs. The \textbf{expansion stage} then introduces caption-augmented configurations: $IC$ and $C$, to test whether automatically generated captions enhance retrieval robustness. This sequential experiment ensures any observed improvements that is directly attributed to image caption rather than variance in the foundational stage. In this work, Qwen2-VL-2B-Instruct \cite{Qwen-VL, Qwen2VL} is employed to generate image captions. The prompt is shown in Appendix \ref{caption_prompt}. 

From Table \ref{tab:fou_ret_result}, the $I \leftrightarrow IT$ configuration demonstrates superior performance across both datasets when paired with EVA-CLIP$_{SF}$ \cite{evaclip-1,evaclip-2}. This suggests EVA-CLIP’s pretrained vision-language alignment excels at bridging pure image queries with image + text candidates in the knowledge base, and thus motivates our selection of EVA-CLIP as the default retriever in this study. 

In the expansion stage from Table \ref{tab:exp_ret_result}, we observe that augmenting image queries with generated caption ($IC$) yields modest improvements over raw image queries ($I$). This shows that image captions provide complementary semantic signals that enhance the query. However, applying captions to both query and candidates, $IC \leftrightarrow IC$, degrades performance drastically even compared to $I \leftrightarrow I$. With the low retrieval accuracy of $C \leftrightarrow C$, this is likely due to caption discrepancies between query and candidates that amplify visual differences. 
\vspace{-3pt}
\begin{tcolorbox}
\textbf{\textcolor{blue}{Takeaway:}} It is clear that a large-scale CLIP model (in this study, EVA-CLIP) is a robust zero-shot retriever. Furthermore, augmenting image \textbf{on the query side} with generated captions improves Recall@1 accuracy by 1\% over image-only.
\end{tcolorbox}

\section{Re-ranking}
\label{sec:reranking}
While relevant candidates may appear in the top-$K$ retrieval results, modern LVLMs still exhibit a positional attention bias, called the "lost-in-the-middle" effect \cite{litm}. This effect persists even when correct articles are retrieved but present in the middle, as LVLMs disproportionately focus on candidates in the beginning during answer generation. To mitigate this issue, re-ranking aims at pushing the most relevant candidate to the beginning, aligning with LVLMs’ inherent attention patterns. This step is critical because LVLMs’ performance on knowledge-intensive tasks degrades sharply when key information appears later in the input sequence.

\subsection{Experimental Setup}
Following prior work \cite{llm_reranker}, we evaluate three re-ranking approaches. \textbf{Pointwise Ranking} computes absolute relevance scores for individual query-candidate pairs and sorts the scores. \textbf{Pairwise Ranking} compares candidate pairs through relative preference judgments, asking models to select the more relevant option for each query. \textbf{Listwise Ranking} operates on full candidate lists, requiring models to holistically assess and reorder all retrieved items simultaneously.

In this section, the query modality is fixed to image + question ($IQ$) and the candidate modality is image + text ($IT$). We utilize MM-Embed \cite{mm-embed}, Q-Former from EchoSight \cite{echosight}, and Qwen2-VL-7B-Instruct \cite{Qwen-VL, Qwen2VL} as the re-ranker, and take the top-5 retrieval candidates from Section 3 to measure Recall@1 and MRR improvements after re-ranking. Note that the re-ranker from EchoSight is specifically fine-tuned on InfoSeek and E-VQA, while MM-Embed and Qwen2-VL-7B-Instruct are zero-shot re-rankers. This controlled setup isolates the effectiveness of re-ranking from initial retrieval quality. For pointwise ranking, we compute absolute relevance scores by extracting the last-layer embeddings of both queries and candidates from MM-Embed and EchoSight, and then calculating the dot product similarity score. For pairwise and listwise ranking, we prompt Qwen2-VL-7B-Instruct to re-rank. The prompt template is shown in Appendix \ref{rerank_prompt}. 

\subsection{Results}
Table \ref{tab:rerank_result} presents a comparison of re-ranking strategies, focusing on their performance on Recall@1 and MRR for both the InfoSeek and E-VQA datasets. The baseline (w/o re-ranking) establishes a starting point, with Recall@1 of 64.44 and 62.61 and MRR of 0.71 and 0.694 for InfoSeek and E-VQA, respectively.


For InfoSeek, the zero-shot MM-Embed re-ranker and pairwise ranking degrade performance substantially compared to the baseline. In contrast, EchoSight's fine-tuned re-ranker achieves near-baseline Recall@1 and improves MRR by 0.011 becasue of the fine-tuning on both knowledge bases. Surprisingly, listwise ranking with QwenVL surpasses EchoSight’s performance, demonstrating that LVLMs are inherently a good re-ranker.

Similarly, for E-VQA, EchoSight's fine-tuned re-ranker outperforms listwise ranking but requires training on both knowledge bases, while listwise offers competitive zero-shot performance without fine-tuning. Our experimental findings are consistent with previous work \cite{llm_good_reranker}, indicating that while LVLMs are not effective for initial retrieval, they are good when used for re-ranking retrieved candidates in a zero-shot manner.

\begin{table}[t]
\caption{Re-ranking results across different strategies and datasets. The baseline, w/o Re-rank, is the original retrieval result with $IC \leftrightarrow IT$ from Section \ref{ret_section}.} 
\label{tab:rerank_result}
\begin{adjustbox}{width=210pt}
\begin{tabular}{c|c|lll}
\specialrule{.2em}{.05em}{.05em} 
\multirow{2}{*}{\textbf{Dataset}}  & \multirow{2}{*}{\textbf{Strategy}} & \multicolumn{2}{c}{\textbf{Metric}}   \\ \cline{3-4}  
                                   &  & \multicolumn{1}{c|}{Recall@1 $\uparrow$}    & \multicolumn{1}{c}{MRR $\uparrow$}   \\ 
\specialrule{.1em}{.05em}{.05em}
\multirow{5}{*}{InfoSeek} & w/o Re-rank  & \multicolumn{1}{l|}{64.44$_{+0}$} & \multicolumn{1}{l}{0.71$_{+0}$}   \\
                          & MM-Embed  & \multicolumn{1}{l|}{33.18$_{\textcolor{red}{-31.26}}$} & \multicolumn{1}{l}{0.47$_{\textcolor{red}{-0.24}}$}   \\  
                          & EchoSight & \multicolumn{1}{l|}{64.40$_{\textcolor{red}{-0.04}}$}  & \multicolumn{1}{l}{0.72$_{\textcolor{teal}{+0.01}}$}   \\ 
                          & Pairwise & \multicolumn{1}{l|}{51.84$_{\textcolor{red}{-12.6}}$} & \multicolumn{1}{l}{0.63$_{\textcolor{red}{-0.08}}$}   \\
                          & Listwise & \multicolumn{1}{l|}{65.88$_{\textcolor{teal}{+1.44}}$} & \multicolumn{1}{l}{0.73$_{\textcolor{teal}{+0.02}}$}   \\                
\hline
\multirow{5}{*}{E-VQA}    & w/o Re-rank  & \multicolumn{1}{l|}{62.61$_{+0}$} & \multicolumn{1}{l}{0.69$_{+0}$}   \\
                          & MM-Embed  & \multicolumn{1}{l|}{43.85$_{\textcolor{red}{-18.76}}$} & \multicolumn{1}{l}{0.56$_{\textcolor{red}{-0.13}}$}   \\  
                          & EchoSight & \multicolumn{1}{l|}{69.81$_{\textcolor{teal}{+7.2}}$}  & \multicolumn{1}{l}{0.74$_{\textcolor{teal}{+0.05}}$}   \\ 
                          & Pairwise & \multicolumn{1}{l|}{56.34$_{\textcolor{red}{-6.27}}$} & \multicolumn{1}{l}{0.67$_{\textcolor{red}{-0.02}}$}   \\
                          & Listwise & \multicolumn{1}{l|}{66.42$_{\textcolor{teal}{+3.81}}$} & \multicolumn{1}{l}{0.72$_{\textcolor{teal}{+0.03}}$}   \\     
                          
\specialrule{.2em}{.05em}{.05em} 
\end{tabular}
\end{adjustbox}
\end{table}

\begin{tcolorbox}
\textbf{\textcolor{blue}{Takeaway:}} Listwise ranking with LVLMs is an effective zero-shot re-ranking strategy, may even surpass the performance of fine-tuned re-rankers and yielding an average of 2.6\% improvement in Recall@1 accuracy over two datasets.
\end{tcolorbox}
\section{Generation}
\label{gen_section}
The generation phase synthesizes retrieved knowledge into accurate, contextually grounded answers, where retrieval quality may directly impact answer correctness. This section evaluates the generation capabilities and explores how retrieval influences the quality of the generated responses. We evaluate four conditions. \textbf{Generation without retrieval} answers the question solely without any knowledge provided, serving as the lower bound. \textbf{Generation with initial retrieval} synthesizes responses using top-K retrieved documents before re-ranking. \textbf{Generation after re-ranking} uses optimized candidate ordering to enhance answer accuracy by aligning with VLMs’ positional attention biases. \textbf{Generation with gold document} takes the document containing the answer to the query as a reference to measure the upper bound accuracy.

\begin{figure*}[!ht]
  \includegraphics[width=0.99\textwidth]{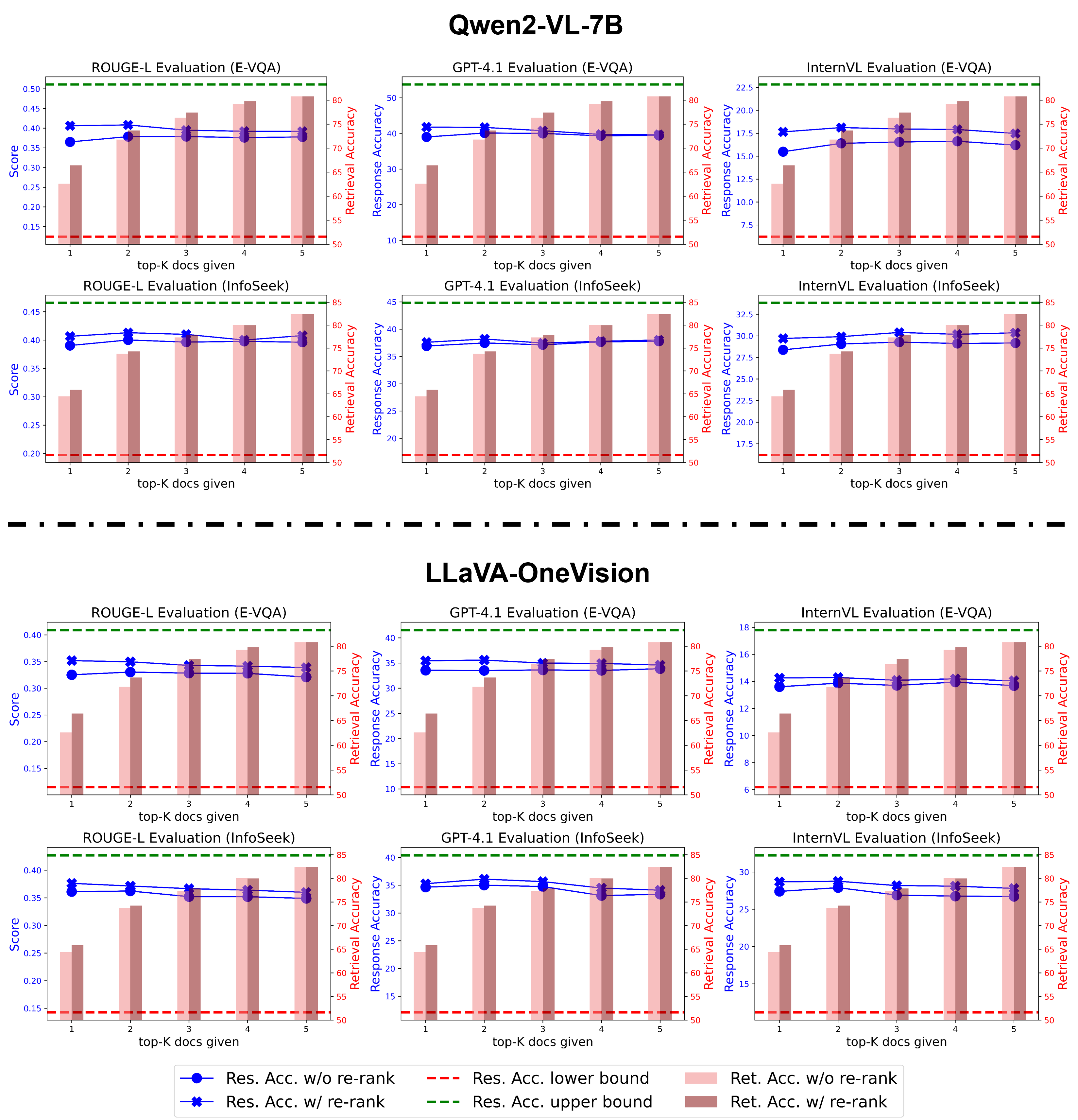}
  \caption{Generation performance of Qwen2-VL-7B-Instruct (top) and LLaVA-OneVision (bottom) across different evaluation metrics on E-VQA and InfoSeek. The line plots correspond to the left y-axis, while the bar plots correspond to the right y-axis. The \textcolor{OliveGreen}{green dashed line} marks the performance of \textbf{Generation with gold document} and the \textcolor{red}{red dashed line} marks the \textbf{Generation without retrieval} of achievable response accuracy. The left y-axis, shown in blue, is the ROUGE-L score and Response Accuracy (Res. Acc.) for \textbf{Generation with initial retrieval} and \textbf{Generation after re-ranking}. The right y-axis, in red, represents the Retrieval Accuracy (Ret. Acc.) as the number of top-K retrieved documents varies.}
  \label{fig:gen_response}
\end{figure*}

\subsection{Experimental Setup}
We evaluate two state-of-the-art LVLMs: Qwen2-VL-7B-Instruct \cite{Qwen2VL} and LLaVA-OneVision \cite{llavanext} to assess how retrieval information improves answer quality. Both models operate in zero-shot mode, leveraging their pretrained multimodal understanding without task-specific fine-tuning.
To assess answer quality, we compute ROUGE-L \cite{rouge} score against reference answers. However, this traditional metric may not sufficiently capture semantic meanings. As a result, we also employ InternVL3-14B \cite{internvl, internvl3} and GPT-4.1 \cite{mm-vet, vlmevalkit, gpt4} as automated judges. The judge receives the generated answer and the reference answer, then checks if the generated answer is correct (answer aligns with reference answer) or incorrect (factually wrong or irrelevant). The prompt template for the judge is shown in Appendix \ref{generation_prompts}.

\subsection{Results}
Figure \ref{fig:gen_response} shows the response accuracy with Qwen2-VL-7B-Instruct. The results demonstrate a critical divergence between retrieval accuracy and response accuracy. While retrieval accuracy (Ret. Acc.) monotonically increases with larger $K$, response accuracy (Res. Acc.) does not improve and even declines. For instance, in E-VQA after re-ranking, Ret. Acc. at top-1 achieves 66.42\% and increases to 80.8\% at top-5, with 14.38\% improvement. However, ROUGE-L score drops from 0.416 to 0.392 and Res. Acc. decreases 0.17\% and 2.11\% with InternVL3 and GPT 4.1 evaluation, correspondingly.

A similar trend is also observed at the bottom of Figure \ref{fig:gen_response}. Our experiments suggest that LVLMs indeed exhibit a strong positional bias, prioritizing information from the initial positions of the input context, so adding more documents may make LVLMs overlook key details or be confused by the irrelevant documents. Thus, re-ranking is necessary to push the most relevant to the beginning.

\begin{tcolorbox}
\textbf{\textcolor{blue}{Takeaway:}}  While re-ranking retrieved results boosts generation accuracy by at least 1\%, adding more documents does not improve generation accuracy, even if the correct answer is present among them. Thus, providing only the most relevant document as a reference is optimal.
\end{tcolorbox}
\section{Unifying Re-ranking and Generation}
Our experiments in Section \ref{gen_section} demonstrate that adding less-relevant documents may not benefit response accuracy, motivating our attempt to explore the potential of unifying re-ranking and generation into a single agentic framework. This approach incorporates a self-reflection loop where the model evaluates both the query and retrieved documents through multiple iterations, and decides the most relevant document. Unlike prior methods like \cite{rankrag}, which rely on instruction-tuned and text-only LLMs, we explore the possibility of LVLMs to dynamically assess query-document relevance and prioritize critical evidence without specific training.

\begin{figure*}
  \includegraphics[width=\textwidth]{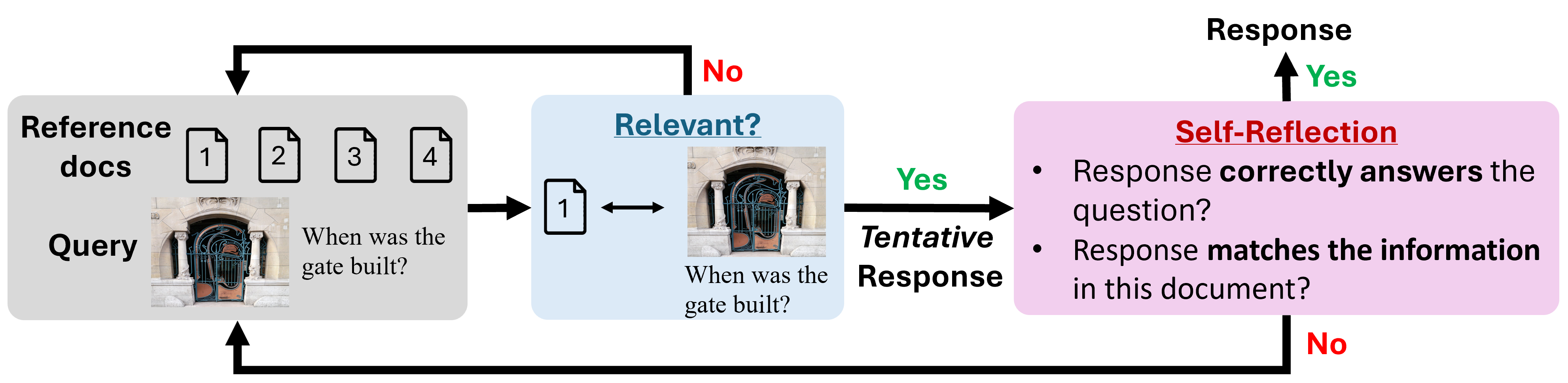}
  \vspace{-2mm}
  \caption{The self-reflection process of unifying re-ranking and generation in a single agentic framework.}
  \vspace{-2mm}
  
  \label{self-reflect}
\end{figure*}

\subsection{Experimental Setup}
Similar to Section \ref{gen_section}, we employ Qwen2-VL-7B \cite{Qwen2VL} and LLaVA-OneVision \cite{llavanext} as the base LVLMs. At each iteration, the model assesses whether the current document contains evidence directly addressing the query. If a document is relevant, the model generates a tentative answer and checks its validity against the document’s content via a self-reflection prompt. A valid response is returned, while an invalid response prompts the model to consider the next document in the retrieved set. If none of the documents provide relevant information, the model returns “Model fails to answer the question” and ends the process. Figure \ref{self-reflect} depicts the pipeline of the unification of re-ranking and generation. The prompts are shown in Appendix \ref{agentic_prompt}. To evaluate the performance, we take the ROUGE-L score and response accuracy with top-1 document given after re-ranking from Section \ref{gen_section} as the baseline, named \textbf{non-Unified}, and also employ InternVL3 and GPT 4.1 as automated judges to assess whether the response is semantically \textit{Correct} or \textit{Incorrect}.

\subsection{Results}
From Table \ref{tab:evqa_unified} and \ref{tab:infoseek_unified}, we observe that the unified agent consistently outperforms decoupled re-ranking and generation pipelines across both datasets and LVLM architectures. This demonstrates that integrating self-reflection capabilities directly into the generation process enables LVLMs to validate response relevance against retrieved documents and the query. By iteratively filtering irrelevant evidence while prioritizing critical information through document-level attention, positional bias is avoided in the decoupled approach.

\begin{table}[t]
\centering
\caption{Response accuracy on the \textbf{E-VQA} dataset when unifying re-ranking and generation, compared to separate approaches, across two LVLMs.} 
  \vspace{-2mm}

\label{tab:evqa_unified}
\begin{adjustbox}{width=220pt}
\begin{tabular}{c|c|lll}
\specialrule{.2em}{.05em}{.05em} 
\multirow{2}{*}{\textbf{Model}}  & \multirow{2}{*}{\textbf{Strategy}} & \multicolumn{3}{c}{\textbf{Evaluation Method}}   \\ \cline{3-5}  
                                   &  & \multicolumn{1}{c|}{ROUGE-L}    & \multicolumn{1}{c|}{GPT}  & \multicolumn{1}{c}{InternVL}  \\ 
\specialrule{.1em}{.05em}{.05em}
\multirow{2}{*}{Qwen} & non-Unified  & \multicolumn{1}{c|}{0.41} & \multicolumn{1}{c|}{41.77} & \multicolumn{1}{c}{17.65}   \\
                          & Unified  & \multicolumn{1}{c|}{\textbf{0.43}} & \multicolumn{1}{c|}{\textbf{45.66}} & \multicolumn{1}{c}{\textbf{19.49}}   \\                 
\hline
\multirow{2}{*}{LLaVA} & non-Unified  & \multicolumn{1}{c|}{0.35} & \multicolumn{1}{c|}{35.44} & \multicolumn{1}{c}{14.25}   \\
                          & Unified  & \multicolumn{1}{c|}{\textbf{0.37}} & \multicolumn{1}{c|}{\textbf{40.69}} & \multicolumn{1}{c}{\textbf{16.57}}   \\     
                          
\specialrule{.2em}{.05em}{.05em} 
\end{tabular}
  \vspace{-2mm}

\end{adjustbox}
\end{table}

\begin{table}[t]
\caption{Response accuracy on the \textbf{InfoSeek} dataset.} 
  \vspace{-2mm}

\label{tab:infoseek_unified}
\begin{adjustbox}{width=220pt}
\begin{tabular}{c|c|lll}
\specialrule{.2em}{.05em}{.05em} 
\multirow{2}{*}{\textbf{Model}}  & \multirow{2}{*}{\textbf{Strategy}} & \multicolumn{3}{c}{\textbf{Evaluation Method}}   \\ \cline{3-5}  
                                   &  & \multicolumn{1}{c|}{ROUGE-L}    & \multicolumn{1}{c|}{GPT}  & \multicolumn{1}{c}{InternVL}  \\ 
\specialrule{.1em}{.05em}{.05em}
\multirow{2}{*}{Qwen} & non-Unified  & \multicolumn{1}{c|}{0.41} & \multicolumn{1}{c|}{37.6} & \multicolumn{1}{c}{29.7}   \\
                          & Unified  & \multicolumn{1}{c|}{\textbf{0.42}} & \multicolumn{1}{c|}{\textbf{39.5}} & \multicolumn{1}{c}{\textbf{31}}   \\                 
\hline
\multirow{2}{*}{LLaVA} & non-Unified  & \multicolumn{1}{c|}{0.38} & \multicolumn{1}{c|}{35.28} & \multicolumn{1}{c}{28.68}   \\
                          & Unified  & \multicolumn{1}{c|}{\textbf{0.39}} & \multicolumn{1}{c|}{\textbf{37.86}} & \multicolumn{1}{c}{\textbf{29.72}}   \\     
                          
\specialrule{.2em}{.05em}{.05em} 
\end{tabular}
  \vspace{-2mm}

\end{adjustbox}
\end{table}

\begin{tcolorbox}
\textbf{\textcolor{blue}{Takeaway:}}  The unified agentic framework outperforms decoupled pipelines, boosting the response accuracy by 5\%/2\% for E-VQA/InfoSeek through LVLM’s iterative self-reflection.
\end{tcolorbox}

\section{Conclusion}
In this paper, we systematically revisited the mRAG pipeline, focusing on zero-shot settings for LVLMs. Our study dissected the retrieval phase, revealing that large-scale CLIP models are highly effective as zero-shot retrievers, and that augmenting image queries with generated captions can provide modest gains. We further analyzed re-ranking strategies and found that listwise re-ranking with LVLMs offers strong zero-shot performance. Our generation experiments demonstrated that candidate ordering has a direct impact on answer accuracy, with re-ranking being essential to ensure relevant evidence is prioritized. However, we observed that adding less-relevant documents is not beneficial. As a result, we introduced an unified agentic framework that integrates re-ranking and generation via self-reflection, enabling LVLMs to dynamically filter irrelevant context and enhance answer accuracy without task-specific fine-tuning.

\section*{Limitations}
Although our systematic study provides several suggestions on each phase in mRAG pipeline, several limitations should be mentioned. First, our evaluation is conducted in a zero-shot setting using frozen pre-trained models, which may not fully capture the performance upper bound achievable with task-specific fine-tuning and the model may provide hallucainated responses. Second, the reliance on distilled datasets, while necessary for computational feasibility in this work, could introduce distributional biases that do not entirely reflect real-world scenarios with larger and more diverse knowledge bases. Third, while LVLM-based judges provide scalable evaluation, they may not perfectly align with human judgment, especially for nuanced or open-ended questions. Future work focusing on improving multi-modal alignment and developing human-centered evaluation frameworks for mRAG remains to be explored.

\bibliography{custom}

\begin{thebibliography}{69}
\expandafter\ifx\csname natexlab\endcsname\relax\def\natexlab#1{#1}\fi

\bibitem[{Abootorabi et~al.(2025)Abootorabi, Zobeiri, Dehghani, Mohammadkhani, Mohammadi, Ghahroodi, Baghshah, and Asgari}]{abootorabi2025ask}
Mohammad~Mahdi Abootorabi, Amirhosein Zobeiri, Mahdi Dehghani, Mohammadali Mohammadkhani, Bardia Mohammadi, Omid Ghahroodi, Mahdieh~Soleymani Baghshah, and Ehsaneddin Asgari. 2025.
\newblock Ask in any modality: A comprehensive survey on multimodal retrieval-augmented generation.
\newblock \emph{arXiv preprint arXiv:2502.08826}.

\bibitem[{Achiam et~al.(2023)Achiam, Adler, Agarwal, Ahmad, Akkaya, Aleman, Almeida, Altenschmidt, Altman, Anadkat et~al.}]{gpt4}
Josh Achiam, Steven Adler, Sandhini Agarwal, Lama Ahmad, Ilge Akkaya, Florencia~Leoni Aleman, Diogo Almeida, Janko Altenschmidt, Sam Altman, Shyamal Anadkat, et~al. 2023.
\newblock Gpt-4 technical report.
\newblock \emph{arXiv preprint arXiv:2303.08774}.

\bibitem[{Alonso et~al.(2025)Alonso, Salaberria, Azkune, Barnes, and de~Lacalle}]{vlm_align_1}
I{\~n}igo Alonso, Ander Salaberria, Gorka Azkune, Jeremy Barnes, and Oier~Lopez de~Lacalle. 2025.
\newblock Vision-language models struggle to align entities across modalities.
\newblock \emph{arXiv preprint arXiv:2503.03854}.

\bibitem[{Bai et~al.(2023)Bai, Bai, Yang, Wang, Tan, Wang, Lin, Zhou, and Zhou}]{Qwen-VL}
Jinze Bai, Shuai Bai, Shusheng Yang, Shijie Wang, Sinan Tan, Peng Wang, Junyang Lin, Chang Zhou, and Jingren Zhou. 2023.
\newblock Qwen-vl: A versatile vision-language model for understanding, localization, text reading, and beyond.
\newblock \emph{arXiv preprint arXiv:2308.12966}.

\bibitem[{Bai et~al.(2024)Bai, Wang, Xiao, He, Han, Zhang, and Shou}]{vlm_hallucination_4}
Zechen Bai, Pichao Wang, Tianjun Xiao, Tong He, Zongbo Han, Zheng Zhang, and Mike~Zheng Shou. 2024.
\newblock Hallucination of multimodal large language models: A survey.
\newblock \emph{arXiv preprint arXiv:2404.18930}.

\bibitem[{Caffagni et~al.(2024)Caffagni, Cocchi, Moratelli, Sarto, Cornia, Baraldi, and Cucchiara}]{wiki-llava}
Davide Caffagni, Federico Cocchi, Nicholas Moratelli, Sara Sarto, Marcella Cornia, Lorenzo Baraldi, and Rita Cucchiara. 2024.
\newblock Wiki-llava: Hierarchical retrieval-augmented generation for multimodal llms.
\newblock In \emph{Proceedings of the IEEE/CVF Conference on Computer Vision and Pattern Recognition}, pages 1818--1826.

\bibitem[{Cekinel et~al.(2025)Cekinel, Karagoz, and {\c{C}}{\"o}ltekin}]{vlm_verify_3}
Recep~Firat Cekinel, Pinar Karagoz, and {\c{C}}a{\u{g}}r{\i} {\c{C}}{\"o}ltekin. 2025.
\newblock \href {https://aclanthology.org/2025.coling-main.310/} {Multimodal fact-checking with vision language models: A probing classifier based solution with embedding strategies}.
\newblock In \emph{Proceedings of the 31st International Conference on Computational Linguistics}, pages 4622--4633, Abu Dhabi, UAE. Association for Computational Linguistics.

\bibitem[{Chen et~al.(2024{\natexlab{a}})Chen, Xu, Kirmani, Ichter, Sadigh, Guibas, and Xia}]{spatialvlm}
Boyuan Chen, Zhuo Xu, Sean Kirmani, Brain Ichter, Dorsa Sadigh, Leonidas Guibas, and Fei Xia. 2024{\natexlab{a}}.
\newblock Spatialvlm: Endowing vision-language models with spatial reasoning capabilities.
\newblock In \emph{Proceedings of the IEEE/CVF Conference on Computer Vision and Pattern Recognition}, pages 14455--14465.

\bibitem[{Chen et~al.(2024{\natexlab{b}})Chen, Lin, Han, and Sun}]{rag3}
Jiawei Chen, Hongyu Lin, Xianpei Han, and Le~Sun. 2024{\natexlab{b}}.
\newblock Benchmarking large language models in retrieval-augmented generation.
\newblock In \emph{Proceedings of the AAAI Conference on Artificial Intelligence}, volume~38, pages 17754--17762.

\bibitem[{Chen et~al.(2023)Chen, Hu, Luan, Sun, Changpinyo, Ritter, and Chang}]{infoseek}
Yang Chen, Hexiang Hu, Yi~Luan, Haitian Sun, Soravit Changpinyo, Alan Ritter, and Ming-Wei Chang. 2023.
\newblock \href {https://doi.org/10.18653/v1/2023.emnlp-main.925} {Can pre-trained vision and language models answer visual information-seeking questions?}
\newblock In \emph{Proceedings of the 2023 Conference on Empirical Methods in Natural Language Processing}, pages 14948--14968, Singapore. Association for Computational Linguistics.

\bibitem[{Chen et~al.(2024{\natexlab{c}})Chen, Wu, Wang, Su, Chen, Xing, Zhong, Zhang, Zhu, Lu et~al.}]{internvl}
Zhe Chen, Jiannan Wu, Wenhai Wang, Weijie Su, Guo Chen, Sen Xing, Muyan Zhong, Qinglong Zhang, Xizhou Zhu, Lewei Lu, et~al. 2024{\natexlab{c}}.
\newblock Internvl: Scaling up vision foundation models and aligning for generic visual-linguistic tasks.
\newblock In \emph{Proceedings of the IEEE/CVF Conference on Computer Vision and Pattern Recognition}, pages 24185--24198.

\bibitem[{Chow et~al.(2025)Chow, Mao, Li, Seita, Guizilini, and Wang}]{phys_vlm_2}
Wei Chow, Jiageng Mao, Boyi Li, Daniel Seita, Vitor Guizilini, and Yue Wang. 2025.
\newblock Physbench: Benchmarking and enhancing vision-language models for physical world understanding.
\newblock \emph{arXiv preprint arXiv:2501.16411}.

\bibitem[{Douze et~al.(2024)Douze, Guzhva, Deng, Johnson, Szilvasy, Mazar{\'e}, Lomeli, Hosseini, and J{\'e}gou}]{faiss}
Matthijs Douze, Alexandr Guzhva, Chengqi Deng, Jeff Johnson, Gergely Szilvasy, Pierre-Emmanuel Mazar{\'e}, Maria Lomeli, Lucas Hosseini, and Herv{\'e} J{\'e}gou. 2024.
\newblock The faiss library.
\newblock \emph{arXiv preprint arXiv:2401.08281}.

\bibitem[{Duan et~al.(2024)Duan, Yang, Qiao, Fang, Chen, Liu, Dong, Zang, Zhang, Wang et~al.}]{vlmevalkit}
Haodong Duan, Junming Yang, Yuxuan Qiao, Xinyu Fang, Lin Chen, Yuan Liu, Xiaoyi Dong, Yuhang Zang, Pan Zhang, Jiaqi Wang, et~al. 2024.
\newblock Vlmevalkit: An open-source toolkit for evaluating large multi-modality models.
\newblock In \emph{Proceedings of the 32nd ACM international conference on multimedia}, pages 11198--11201.

\bibitem[{Favero et~al.(2024)Favero, Zancato, Trager, Choudhary, Perera, Achille, Swaminathan, and Soatto}]{vlm_hallucination_1}
Alessandro Favero, Luca Zancato, Matthew Trager, Siddharth Choudhary, Pramuditha Perera, Alessandro Achille, Ashwin Swaminathan, and Stefano Soatto. 2024.
\newblock Multi-modal hallucination control by visual information grounding.
\newblock In \emph{Proceedings of the IEEE/CVF Conference on Computer Vision and Pattern Recognition}, pages 14303--14312.

\bibitem[{Gangi~Reddy et~al.(2024)Gangi~Reddy, Doo, Xu, Sultan, Swain, Sil, and Ji}]{first}
Revanth Gangi~Reddy, JaeHyeok Doo, Yifei Xu, Md~Arafat Sultan, Deevya Swain, Avirup Sil, and Heng Ji. 2024.
\newblock \href {https://doi.org/10.18653/v1/2024.emnlp-main.491} {{FIRST}: Faster improved listwise reranking with single token decoding}.
\newblock In \emph{Proceedings of the 2024 Conference on Empirical Methods in Natural Language Processing}, pages 8642--8652, Miami, Florida, USA. Association for Computational Linguistics.

\bibitem[{Gao et~al.(2024)Gao, Sarkar, Xia, Xiao, Wu, Ichter, Majumdar, and Sadigh}]{phys_vlm_3}
Jensen Gao, Bidipta Sarkar, Fei Xia, Ted Xiao, Jiajun Wu, Brian Ichter, Anirudha Majumdar, and Dorsa Sadigh. 2024.
\newblock Physically grounded vision-language models for robotic manipulation.
\newblock In \emph{2024 IEEE International Conference on Robotics and Automation (ICRA)}, pages 12462--12469. IEEE.

\bibitem[{Gao et~al.(2023)Gao, Xiong, Gao, Jia, Pan, Bi, Dai, Sun, Wang, and Wang}]{rag2}
Yunfan Gao, Yun Xiong, Xinyu Gao, Kangxiang Jia, Jinliu Pan, Yuxi Bi, Yixin Dai, Jiawei Sun, Haofen Wang, and Haofen Wang. 2023.
\newblock Retrieval-augmented generation for large language models: A survey.
\newblock \emph{arXiv preprint arXiv:2312.10997}, 2:1.

\bibitem[{Jiang et~al.(2024)Jiang, Song, Zhang, Huang, Deng, Sun, Zhang, Wang, and Zhuang}]{e5-v}
Ting Jiang, Minghui Song, Zihan Zhang, Haizhen Huang, Weiwei Deng, Feng Sun, Qi~Zhang, Deqing Wang, and Fuzhen Zhuang. 2024.
\newblock E5-v: Universal embeddings with multimodal large language models.
\newblock \emph{arXiv preprint arXiv:2407.12580}.

\bibitem[{Lewis et~al.(2020)Lewis, Perez, Piktus, Petroni, Karpukhin, Goyal, K{\"u}ttler, Lewis, Yih, Rockt{\"a}schel et~al.}]{rag1}
Patrick Lewis, Ethan Perez, Aleksandra Piktus, Fabio Petroni, Vladimir Karpukhin, Naman Goyal, Heinrich K{\"u}ttler, Mike Lewis, Wen-tau Yih, Tim Rockt{\"a}schel, et~al. 2020.
\newblock Retrieval-augmented generation for knowledge-intensive nlp tasks.
\newblock \emph{Advances in neural information processing systems}, 33:9459--9474.

\bibitem[{Li et~al.(2024)Li, Zhang, Guo, Zhang, Li, Zhang, Zhang, Zhang, Li, Liu et~al.}]{llavanext}
Bo~Li, Yuanhan Zhang, Dong Guo, Renrui Zhang, Feng Li, Hao Zhang, Kaichen Zhang, Peiyuan Zhang, Yanwei Li, Ziwei Liu, et~al. 2024.
\newblock Llava-onevision: Easy visual task transfer.
\newblock \emph{arXiv preprint arXiv:2408.03326}.

\bibitem[{Li et~al.(2023)Li, Wu, Abbeel, and Malik}]{vlm_task_3}
Boyi Li, Philipp Wu, Pieter Abbeel, and Jitendra Malik. 2023.
\newblock Interactive task planning with language models.
\newblock \emph{arXiv preprint arXiv:2310.10645}.

\bibitem[{Liao et~al.(2024)Liao, Mahmood, Fidler, and Acuna}]{vlm_semantic_1}
Yuan-Hong Liao, Rafid Mahmood, Sanja Fidler, and David Acuna. 2024.
\newblock Can feedback enhance semantic grounding in large vision-language models?
\newblock \emph{arXiv preprint arXiv:2404.06510}.

\bibitem[{Lin(2004)}]{rouge}
Chin-Yew Lin. 2004.
\newblock \href {https://aclanthology.org/W04-1013/} {{ROUGE}: A package for automatic evaluation of summaries}.
\newblock In \emph{Text Summarization Branches Out}, pages 74--81, Barcelona, Spain. Association for Computational Linguistics.

\bibitem[{Lin et~al.(2024)Lin, Lee, Shoeybi, Lin, Catanzaro, and Ping}]{mm-embed}
Sheng-Chieh Lin, Chankyu Lee, Mohammad Shoeybi, Jimmy Lin, Bryan Catanzaro, and Wei Ping. 2024.
\newblock Mm-embed: Universal multimodal retrieval with multimodal llms.
\newblock \emph{arXiv preprint arXiv:2411.02571}.

\bibitem[{Liu et~al.(2024)Liu, Lin, Hewitt, Paranjape, Bevilacqua, Petroni, and Liang}]{litm}
Nelson~F. Liu, Kevin Lin, John Hewitt, Ashwin Paranjape, Michele Bevilacqua, Fabio Petroni, and Percy Liang. 2024.
\newblock \href {https://doi.org/10.1162/tacl_a_00638} {Lost in the middle: How language models use long contexts}.
\newblock \emph{Transactions of the Association for Computational Linguistics}, 12:157--173.

\bibitem[{Liu et~al.(2025)Liu, Duan, Chen, Lu, Sun, and Mao}]{llm_reranker}
Qi~Liu, Haozhe Duan, Yiqun Chen, Quanfeng Lu, Weiwei Sun, and Jiaxin Mao. 2025.
\newblock Llm4ranking: An easy-to-use framework of utilizing large language models for document reranking.
\newblock \emph{arXiv preprint arXiv:2504.07439}.

\bibitem[{Lu et~al.(2024)Lu, Li, Chen, Xu, Luo, Zhang, and Ye}]{ovis}
Shiyin Lu, Yang Li, Qing-Guo Chen, Zhao Xu, Weihua Luo, Kaifu Zhang, and Han-Jia Ye. 2024.
\newblock Ovis: Structural embedding alignment for multimodal large language model.
\newblock \emph{arXiv preprint arXiv:2405.20797}.

\bibitem[{Ma et~al.(2023)Ma, Cao, Hong, and Sun}]{llm_good_reranker}
Yubo Ma, Yixin Cao, Yong Hong, and Aixin Sun. 2023.
\newblock \href {https://doi.org/10.18653/v1/2023.findings-emnlp.710} {Large language model is not a good few-shot information extractor, but a good reranker for hard samples!}
\newblock In \emph{Findings of the Association for Computational Linguistics: EMNLP 2023}, pages 10572--10601, Singapore. Association for Computational Linguistics.

\bibitem[{Mei et~al.(2025)Mei, Mo, Yang, and Chen}]{mei2025survey}
Lang Mei, Siyu Mo, Zhihan Yang, and Chong Chen. 2025.
\newblock A survey of multimodal retrieval-augmented generation.
\newblock \emph{arXiv preprint arXiv:2504.08748}.

\bibitem[{Mensink et~al.(2023)Mensink, Uijlings, Castrejon, Goel, Cadar, Zhou, Sha, Araujo, and Ferrari}]{evqa}
Thomas Mensink, Jasper Uijlings, Lluis Castrejon, Arushi Goel, Felipe Cadar, Howard Zhou, Fei Sha, Andr{\'e} Araujo, and Vittorio Ferrari. 2023.
\newblock Encyclopedic vqa: Visual questions about detailed properties of fine-grained categories.
\newblock In \emph{Proceedings of the IEEE/CVF International Conference on Computer Vision}, pages 3113--3124.

\bibitem[{Mortaheb et~al.(2025)Mortaheb, Khojastepour, Chakradhar, and Ulukus}]{rerank_score}
Matin Mortaheb, Mohammad A~Amir Khojastepour, Srimat~T Chakradhar, and Sennur Ulukus. 2025.
\newblock Re-ranking the context for multimodal retrieval augmented generation.
\newblock \emph{arXiv preprint arXiv:2501.04695}.

\bibitem[{Prabhu et~al.(2024)Prabhu, Purushwalkam, Yan, Xiong, and Xu}]{vlm_verify_1}
Viraj Prabhu, Senthil Purushwalkam, An~Yan, Caiming Xiong, and Ran Xu. 2024.
\newblock Trust but verify: Programmatic vlm evaluation in the wild.
\newblock \emph{arXiv preprint arXiv:2410.13121}.

\bibitem[{Qin et~al.(2023)Qin, Jagerman, Hui, Zhuang, Wu, Yan, Shen, Liu, Liu, Metzler et~al.}]{pairwise}
Zhen Qin, Rolf Jagerman, Kai Hui, Honglei Zhuang, Junru Wu, Le~Yan, Jiaming Shen, Tianqi Liu, Jialu Liu, Donald Metzler, et~al. 2023.
\newblock Large language models are effective text rankers with pairwise ranking prompting.
\newblock \emph{arXiv preprint arXiv:2306.17563}.

\bibitem[{Rawte et~al.(2025)Rawte, Mishra, Sheth, and Das}]{vlm_hallucination_2}
Vipula Rawte, Aryan Mishra, Amit Sheth, and Amitava Das. 2025.
\newblock Defining and quantifying visual hallucinations in vision-language models.
\newblock In \emph{Proceedings of the 5th Workshop on Trustworthy NLP (TrustNLP 2025)}, pages 501--510.

\bibitem[{Ren et~al.(2025)Ren, Wang, Zhou, Zhao, Wang, Liu, Wen, and Chua}]{listwise}
Ruiyang Ren, Yuhao Wang, Kun Zhou, Wayne~Xin Zhao, Wenjie Wang, Jing Liu, Ji-Rong Wen, and Tat-Seng Chua. 2025.
\newblock Self-calibrated listwise reranking with large language models.
\newblock In \emph{Proceedings of the ACM on Web Conference 2025}, pages 3692--3701.

\bibitem[{Riedler and Langer(2024)}]{riedler2024beyond}
Monica Riedler and Stefan Langer. 2024.
\newblock Beyond text: Optimizing rag with multimodal inputs for industrial applications.
\newblock \emph{arXiv preprint arXiv:2410.21943}.

\bibitem[{Sahoo et~al.(2024)Sahoo, Meharia, Ghosh, Saha, Jain, and Chadha}]{vlm_hallucination_3}
Pranab Sahoo, Prabhash Meharia, Akash Ghosh, Sriparna Saha, Vinija Jain, and Aman Chadha. 2024.
\newblock A comprehensive survey of hallucination in large language, image, video and audio foundation models.
\newblock \emph{arXiv preprint arXiv:2405.09589}.

\bibitem[{Sahu et~al.(2024)Sahu, Sikka, and Divakaran}]{vlm_verify_2}
Pritish Sahu, Karan Sikka, and Ajay Divakaran. 2024.
\newblock \href {https://doi.org/10.18653/v1/2024.emnlp-main.470} {Pelican: Correcting hallucination in vision-{LLM}s via claim decomposition and program of thought verification}.
\newblock In \emph{Proceedings of the 2024 Conference on Empirical Methods in Natural Language Processing}, pages 8228--8248, Miami, Florida, USA. Association for Computational Linguistics.

\bibitem[{Sinha et~al.(2024)Sinha, Jain, and Chadha}]{guiding}
Neelabh Sinha, Vinija Jain, and Aman Chadha. 2024.
\newblock Guiding vision-language model selection for visual question-answering across tasks, domains, and knowledge types.
\newblock \emph{arXiv preprint arXiv:2409.09269}.

\bibitem[{Siyue et~al.(2024)Siyue, Yuxiang, Yiming, Xiaobao, Tuan, and Chen}]{tsqa_1}
Zhang Siyue, Xue Yuxiang, Zhang Yiming, Wu~Xiaobao, Luu~Anh Tuan, and Zhao Chen. 2024.
\newblock Mrag: A modular retrieval framework for time-sensitive question answering.
\newblock \emph{arXiv preprint arXiv:2412.15540}.

\bibitem[{Sun et~al.(2023)Sun, Fang, Wu, Wang, and Cao}]{evaclip-1}
Quan Sun, Yuxin Fang, Ledell Wu, Xinlong Wang, and Yue Cao. 2023.
\newblock Eva-clip: Improved training techniques for clip at scale.
\newblock \emph{arXiv preprint arXiv:2303.15389}.

\bibitem[{Sun et~al.(2024)Sun, Wang, Yu, Cui, Zhang, Zhang, and Wang}]{evaclip-2}
Quan Sun, Jinsheng Wang, Qiying Yu, Yufeng Cui, Fan Zhang, Xiaosong Zhang, and Xinlong Wang. 2024.
\newblock Eva-clip-18b: Scaling clip to 18 billion parameters.
\newblock \emph{arXiv preprint arXiv:2402.04252}.

\bibitem[{Uddin et~al.(2024)Uddin, Saeidi, Handa, Seth, Son, Blanco, Corman, and Baral}]{tsqa_3}
Md~Nayem Uddin, Amir Saeidi, Divij Handa, Agastya Seth, Tran~Cao Son, Eduardo Blanco, Steven~R Corman, and Chitta Baral. 2024.
\newblock Unseentimeqa: Time-sensitive question-answering beyond llms' memorization.
\newblock \emph{arXiv preprint arXiv:2407.03525}.

\bibitem[{Van~Horn et~al.(2021)Van~Horn, Cole, Beery, Wilber, Belongie, and Mac~Aodha}]{inat}
Grant Van~Horn, Elijah Cole, Sara Beery, Kimberly Wilber, Serge Belongie, and Oisin Mac~Aodha. 2021.
\newblock Benchmarking representation learning for natural world image collections.
\newblock In \emph{Proceedings of the IEEE/CVF conference on computer vision and pattern recognition}, pages 12884--12893.

\bibitem[{Wan et~al.(2024)Wan, Cho, Stengel-Eskin, and Bansal}]{vlm_semantic_2}
David Wan, Jaemin Cho, Elias Stengel-Eskin, and Mohit Bansal. 2024.
\newblock Contrastive region guidance: Improving grounding in vision-language models without training.
\newblock In \emph{European Conference on Computer Vision}, pages 198--215. Springer.

\bibitem[{Wang et~al.(2024)Wang, Bai, Tan, Wang, Fan, Bai, Chen, Liu, Wang, Ge, Fan, Dang, Du, Ren, Men, Liu, Zhou, Zhou, and Lin}]{Qwen2VL}
Peng Wang, Shuai Bai, Sinan Tan, Shijie Wang, Zhihao Fan, Jinze Bai, Keqin Chen, Xuejing Liu, Jialin Wang, Wenbin Ge, Yang Fan, Kai Dang, Mengfei Du, Xuancheng Ren, Rui Men, Dayiheng Liu, Chang Zhou, Jingren Zhou, and Junyang Lin. 2024.
\newblock Qwen2-vl: Enhancing vision-language model's perception of the world at any resolution.
\newblock \emph{arXiv preprint arXiv:2409.12191}.

\bibitem[{Wang et~al.(2025)Wang, Kim, Taalimi, Sun, and Kuo}]{vlm_grounding_1}
Shijie Wang, Dahun Kim, Ali Taalimi, Chen Sun, and Weicheng Kuo. 2025.
\newblock Learning visual grounding from generative vision and language model.
\newblock In \emph{2025 IEEE/CVF Winter Conference on Applications of Computer Vision (WACV)}, pages 8057--8067. IEEE.

\bibitem[{Wei et~al.(2024)Wei, Chen, Chen, Hu, Zhang, Fu, Ritter, and Chen}]{uniIR}
Cong Wei, Yang Chen, Haonan Chen, Hexiang Hu, Ge~Zhang, Jie Fu, Alan Ritter, and Wenhu Chen. 2024.
\newblock Uniir: Training and benchmarking universal multimodal information retrievers.
\newblock In \emph{European Conference on Computer Vision}, pages 387--404. Springer.

\bibitem[{Weyand et~al.(2020)Weyand, Araujo, Cao, and Sim}]{gldv2}
Tobias Weyand, Andre Araujo, Bingyi Cao, and Jack Sim. 2020.
\newblock Google landmarks dataset v2-a large-scale benchmark for instance-level recognition and retrieval.
\newblock In \emph{Proceedings of the IEEE/CVF conference on computer vision and pattern recognition}, pages 2575--2584.

\bibitem[{Wu et~al.(2024)Wu, Liu, He, Liu, Zhang, Wang, and Wang}]{tsqa_2}
Feifan Wu, Lingyuan Liu, Wentao He, Ziqi Liu, Zhiqiang Zhang, Haofen Wang, and Meng Wang. 2024.
\newblock Time-sensitve retrieval-augmented generation for question answering.
\newblock In \emph{Proceedings of the 33rd ACM International Conference on Information and Knowledge Management}, pages 2544--2553.

\bibitem[{Xia et~al.(2024{\natexlab{a}})Xia, Zhu, Li, Wang, Shi, Wang, Zhang, Zou, and Yao}]{mmed}
Peng Xia, Kangyu Zhu, Haoran Li, Tianze Wang, Weijia Shi, Sheng Wang, Linjun Zhang, James Zou, and Huaxiu Yao. 2024{\natexlab{a}}.
\newblock Mmed-rag: Versatile multimodal rag system for medical vision language models.
\newblock \emph{arXiv preprint arXiv:2410.13085}.

\bibitem[{Xia et~al.(2024{\natexlab{b}})Xia, Zhu, Li, Zhu, Li, Li, Zhang, and Yao}]{rule}
Peng Xia, Kangyu Zhu, Haoran Li, Hongtu Zhu, Yun Li, Gang Li, Linjun Zhang, and Huaxiu Yao. 2024{\natexlab{b}}.
\newblock Rule: Reliable multimodal rag for factuality in medical vision language models.
\newblock In \emph{Proceedings of the 2024 Conference on Empirical Methods in Natural Language Processing}, pages 1081--1093.

\bibitem[{Xu et~al.(2024)Xu, Huang, Wang, Chen, Pang, and Lin}]{vlm_grounding_2}
Runsen Xu, Zhiwei Huang, Tai Wang, Yilun Chen, Jiangmiao Pang, and Dahua Lin. 2024.
\newblock Vlm-grounder: A vlm agent for zero-shot 3d visual grounding.
\newblock In \emph{CoRL}.

\bibitem[{Yan and Xie(2024)}]{echosight}
Yibin Yan and Weidi Xie. 2024.
\newblock \href {https://aclanthology.org/2024.findings-emnlp.83} {{E}cho{S}ight: Advancing visual-language models with {W}iki knowledge}.
\newblock In \emph{Findings of the Association for Computational Linguistics: EMNLP 2024}, pages 1538--1551, Miami, Florida, USA. Association for Computational Linguistics.

\bibitem[{Yang et~al.(2024)Yang, Garrett, Fox, Lozano-P{\'e}rez, and Kaelbling}]{vlm_task_1}
Zhutian Yang, Caelan Garrett, Dieter Fox, Tom{\'a}s Lozano-P{\'e}rez, and Leslie~Pack Kaelbling. 2024.
\newblock Guiding long-horizon task and motion planning with vision language models.
\newblock \emph{arXiv preprint arXiv:2410.02193}.

\bibitem[{Yang et~al.(2023)Yang, Kafle, Dernoncourt, and Ordonez}]{vlm_grounding_3}
Ziyan Yang, Kushal Kafle, Franck Dernoncourt, and Vicente Ordonez. 2023.
\newblock Improving visual grounding by encouraging consistent gradient-based explanations.
\newblock In \emph{Proceedings of the IEEE/CVF Conference on Computer Vision and Pattern Recognition}, pages 19165--19174.

\bibitem[{Yu et~al.(2023)Yu, Yang, Li, Wang, Lin, Liu, Wang, and Wang}]{mm-vet}
Weihao Yu, Zhengyuan Yang, Linjie Li, Jianfeng Wang, Kevin Lin, Zicheng Liu, Xinchao Wang, and Lijuan Wang. 2023.
\newblock Mm-vet: Evaluating large multimodal models for integrated capabilities.
\newblock \emph{arXiv preprint arXiv:2308.02490}.

\bibitem[{Yu et~al.(2024)Yu, Ping, Liu, Wang, You, Zhang, Shoeybi, and Catanzaro}]{rankrag}
Yue Yu, Wei Ping, Zihan Liu, Boxin Wang, Jiaxuan You, Chao Zhang, Mohammad Shoeybi, and Bryan Catanzaro. 2024.
\newblock Rankrag: Unifying context ranking with retrieval-augmented generation in llms.
\newblock \emph{Advances in Neural Information Processing Systems}, 37:121156--121184.

\bibitem[{Yuan et~al.(2024)Yuan, Sun, Omeiza, Zhao, Newman, Kunze, and Gadd}]{rag-driver}
Jianhao Yuan, Shuyang Sun, Daniel Omeiza, Bo~Zhao, Paul Newman, Lars Kunze, and Matthew Gadd. 2024.
\newblock Rag-driver: Generalisable driving explanations with retrieval-augmented in-context learning in multi-modal large language model.
\newblock \emph{arXiv preprint arXiv:2402.10828}.

\bibitem[{Zhang et~al.(2024{\natexlab{a}})Zhang, Huang, Jin, and Lu}]{vlm4vision}
Jingyi Zhang, Jiaxing Huang, Sheng Jin, and Shijian Lu. 2024{\natexlab{a}}.
\newblock Vision-language models for vision tasks: A survey.
\newblock \emph{IEEE Transactions on Pattern Analysis and Machine Intelligence}.

\bibitem[{Zhang et~al.(2024{\natexlab{b}})Zhang, Luan, Hu, Lee, Qiao, Chen, Su, and Chang}]{MagicLens}
Kai Zhang, Yi~Luan, Hexiang Hu, Kenton Lee, Siyuan Qiao, Wenhu Chen, Yu~Su, and Ming-Wei Chang. 2024{\natexlab{b}}.
\newblock Magiclens: Self-supervised image retrieval with open-ended instructions.
\newblock In \emph{The Forty-first International Conference on Machine Learning (ICML)}.

\bibitem[{Zhang et~al.(2024{\natexlab{c}})Zhang, Zhang, Xie, Li, Dai, Long, Xie, Zhang, Li, and Zhang}]{gme}
Xin Zhang, Yanzhao Zhang, Wen Xie, Mingxin Li, Ziqi Dai, Dingkun Long, Pengjun Xie, Meishan Zhang, Wenjie Li, and Min Zhang. 2024{\natexlab{c}}.
\newblock Gme: Improving universal multimodal retrieval by multimodal llms.
\newblock \emph{arXiv preprint arXiv:2412.16855}.

\bibitem[{Zhaxizhuoma et~al.(2024)Zhaxizhuoma, Chen, Wu, Sun, Wang, Zhou, Cao, Ding, Zhao, and Li}]{vlm_task_2}
Zhaxizhuoma Zhaxizhuoma, Pengan Chen, Ziniu Wu, Jiawei Sun, Dong Wang, Peng Zhou, Nieqing Cao, Yan Ding, Bin Zhao, and Xuelong Li. 2024.
\newblock Alignbot: Aligning vlm-powered customized task planning with user reminders through fine-tuning for household robots.
\newblock \emph{arXiv preprint arXiv:2409.11905}.

\bibitem[{Zhou et~al.(2024)Zhou, Liu, Liu, Xiao, Wang, Zhao, Zhang, Lian, and Xiong}]{bge}
Junjie Zhou, Zheng Liu, Ze~Liu, Shitao Xiao, Yueze Wang, Bo~Zhao, Chen~Jason Zhang, Defu Lian, and Yongping Xiong. 2024.
\newblock Megapairs: Massive data synthesis for universal multimodal retrieval.
\newblock \emph{arXiv preprint arXiv:2412.14475}.

\bibitem[{Zhou et~al.(2025)Zhou, Tao, Zhao, Guo, Dong, Tang, and Wang}]{phys_vlm_1}
Weijie Zhou, Manli Tao, Chaoyang Zhao, Haiyun Guo, Honghui Dong, Ming Tang, and Jinqiao Wang. 2025.
\newblock Physvlm: Enabling visual language models to understand robotic physical reachability.
\newblock \emph{arXiv preprint arXiv:2503.08481}.

\bibitem[{Zhu et~al.(2025)Zhu, Wang, Chen, Liu, Ye, Gu, Duan, Tian, Su, Shao et~al.}]{internvl3}
Jinguo Zhu, Weiyun Wang, Zhe Chen, Zhaoyang Liu, Shenglong Ye, Lixin Gu, Yuchen Duan, Hao Tian, Weijie Su, Jie Shao, et~al. 2025.
\newblock Internvl3: Exploring advanced training and test-time recipes for open-source multimodal models.
\newblock \emph{arXiv preprint arXiv:2504.10479}.

\bibitem[{Zhu et~al.(2024)Zhu, Liu, Wang, Tu, and Chen}]{vlm_align_2}
Tinghui Zhu, Qin Liu, Fei Wang, Zhengzhong Tu, and Muhao Chen. 2024.
\newblock Unraveling cross-modality knowledge conflicts in large vision-language models.
\newblock \emph{arXiv preprint arXiv:2410.03659}.

\bibitem[{Zhuang et~al.(2024)Zhuang, Zhuang, Koopman, and Zuccon}]{rerank_methods}
Shengyao Zhuang, Honglei Zhuang, Bevan Koopman, and Guido Zuccon. 2024.
\newblock A setwise approach for effective and highly efficient zero-shot ranking with large language models.
\newblock In \emph{Proceedings of the 47th International ACM SIGIR Conference on Research and Development in Information Retrieval}, pages 38--47.

\end{thebibliography}
\bibliographystyle{acl_natbib}

\clearpage
\newpage
\appendix
\section{Dataset and Knowledge Base Construction}
In this section, we provide details on the dataset and knowledge base construction.

\subsection{Dataset}
\label{app:dataset_details}
The articles in both E-VQA and InfoSeek are in English, and are about encyclopedic knowledge derived from Wikipedia and Google Landmark datasets. 
In line with previous research \cite{echosight}, we use a total of 4,750 test cases for our evaluation. Our primary focus is on single-hop questions from the original E-VQA dataset \cite{evqa}, ensuring that the model can directly identify the answer by referencing the retrieved documents. For the InfoSeek dataset \cite{infoseek}, since the original release does not include a test set, we sample 5,000 test cases from its validation set to facilitate a fair and consistent evaluation.

\subsection{Knowledge Base}
Given the test cases, we first select articles containing the target answers from the original dataset, then sample additional articles to construct a knowledge base of 50,000 entries. As shown in Table \ref{tab:kb_detail}, we report the approximate construction time and retriever model size for our distilled dataset. For LVLM-based retrievers (BGE-MLLM and GME), we use two NVIDIA RTX A6000 GPUs in parallel with a batch size of 3. For other retrievers, we use a single NVIDIA RTX A6000 GPU with a batch size of 4.  The result demonstrates that LVLM-based retrievers require significantly longer processing time to construct a 50,000-entry knowledge base, which directly motivates our decision to distill the dataset for efficient experimentation.

\begin{table*}[]
\centering
\caption{Details of constructing the knowledge base.} 
\label{tab:kb_detail}
\begin{adjustbox}{width=450pt}
\begin{tabular}{c|cccccc}
\specialrule{.2em}{.05em}{.05em}

                  & CLIP$_{SF}$  & EVA-CLIP$_{SF}$  & BGE-CLIP$_{SF}$  &  BLIP$_{FF}$ & BGE-MLLM & GME \\
\hline \hline
Model Size        & 0.4B  & 7B     & 0.4B   & 2.7B  & 7.57B & 8.2B  \\
\hline
Time - E-VQA (GPU hours)  &  2  &  10.1  &  2   &  4.4   &  39.9  & 40.1 \\ 
\hline
Time - InfoSeek (GPU hours) & 2.1   & 10.4   & 2.1  &  4.7 &  41.4 &  41.7 \\ 
\specialrule{.2em}{.05em}{.05em}        
\end{tabular}
\end{adjustbox}
\end{table*}

\section{Prompt Templates}

\subsection{Image captioning}
\label{caption_prompt}
\begin{figure}[!h]
  \centering
  \includegraphics[width=6cm]{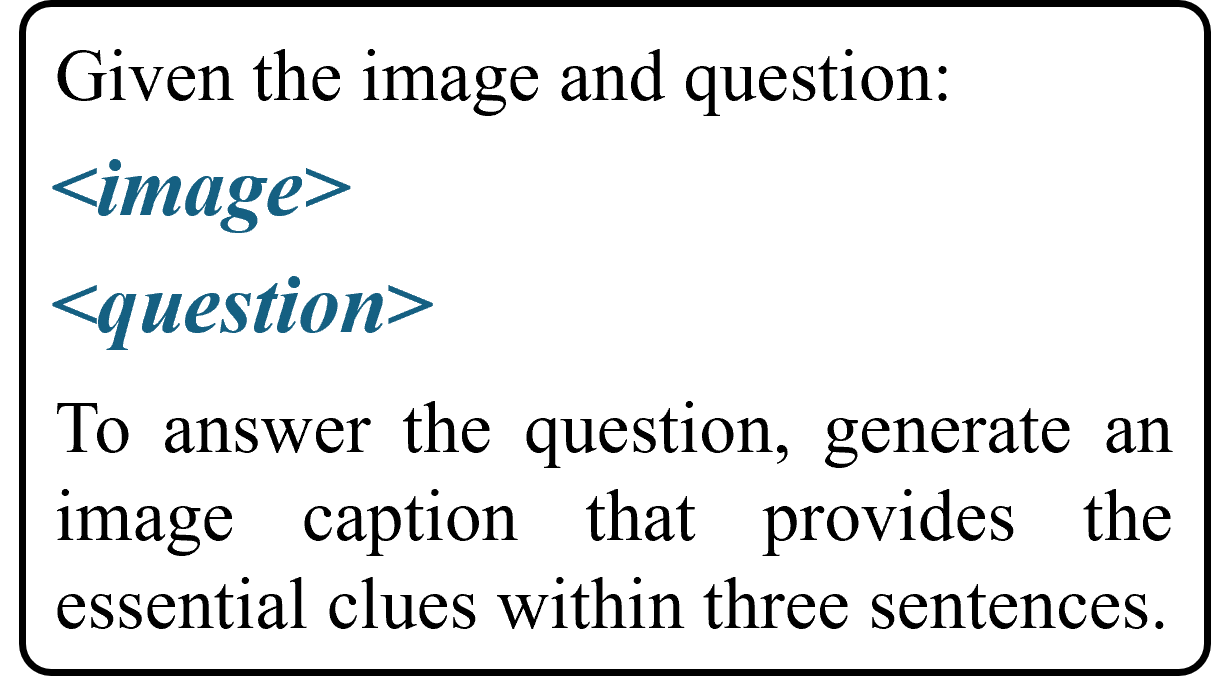}
  \caption{Image captioning prompt on the query side.}
  \label{app:query_cap}
\end{figure}

\begin{figure}[!h]
  \centering
  \includegraphics[width=5cm]{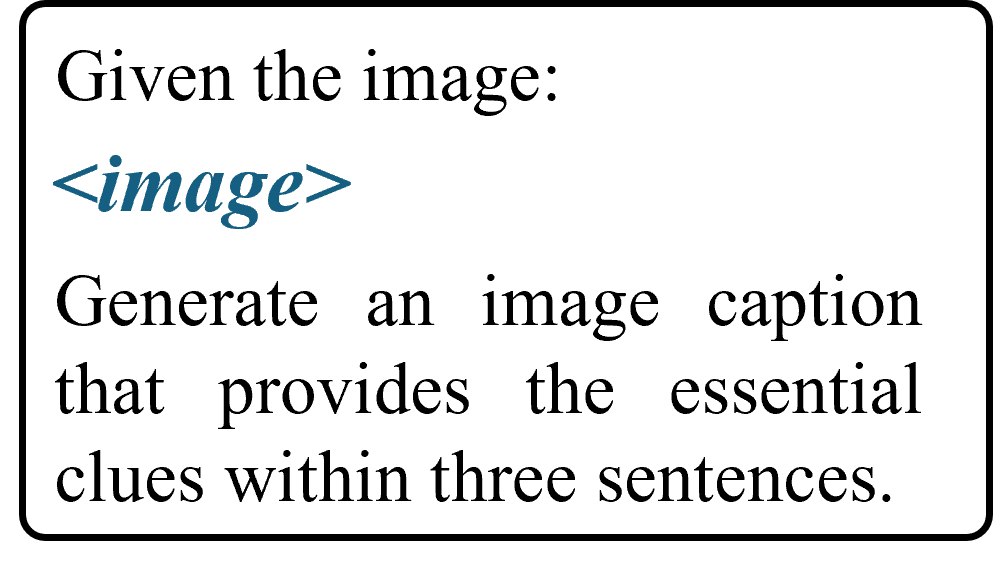}
  \caption{Image captioning prompt on the knowledge base side.}
  \label{app:kb_cap}
\end{figure}

Figure \ref{app:query_cap} presents the prompt to caption the image \textbf{on the query side}. For the knowledge base side, as no question is provided, we caption the image with essential clues only. Figure \ref{app:kb_cap} provides the prompt for captioning images in the knowledge base.

\subsection{Re-ranking}
\label{rerank_prompt}
\begin{figure}[!h]
  \centering
  \includegraphics[width=7.5cm, height=8.5cm]{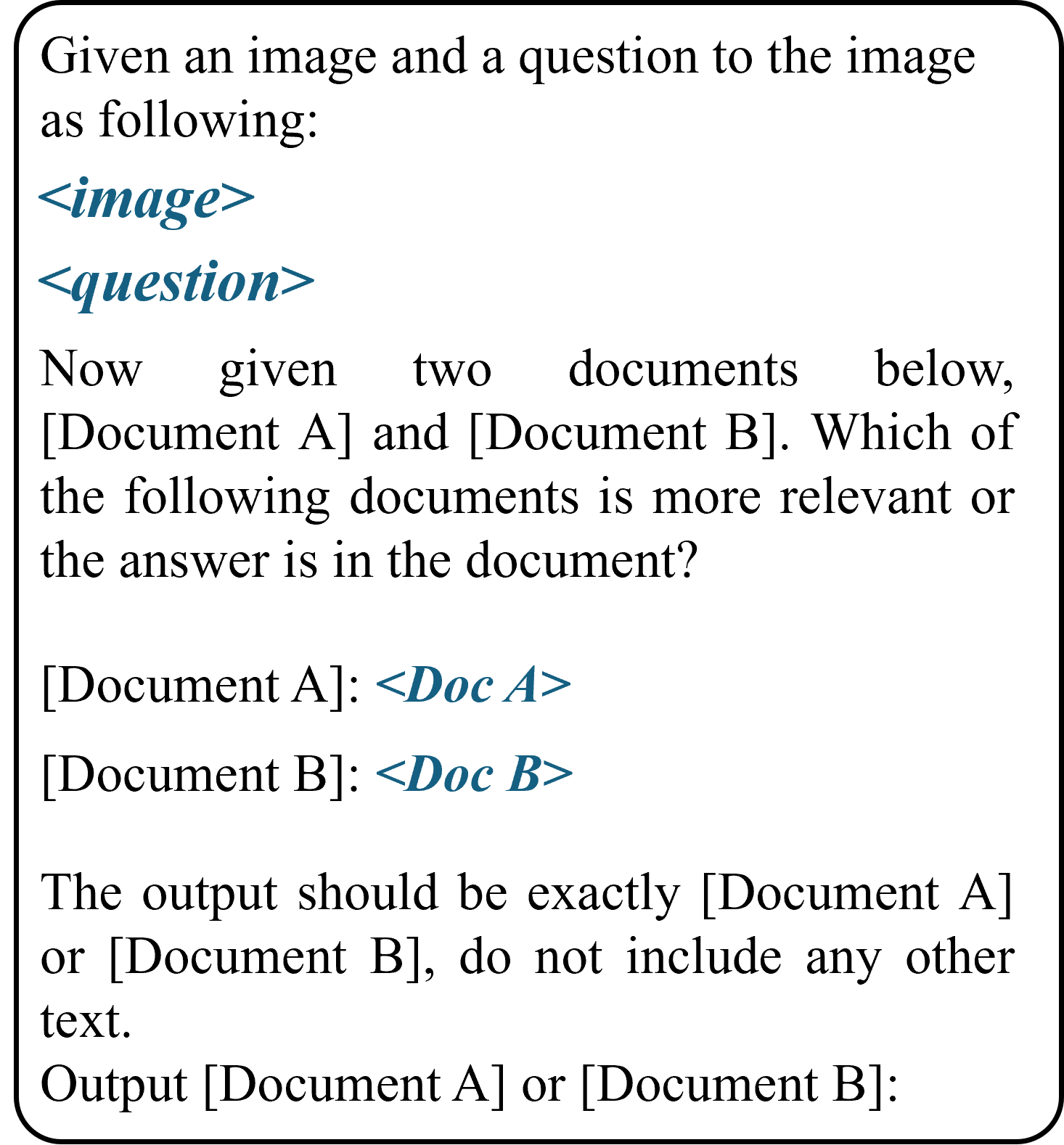}
  \caption{Pairwise re-ranking prompt.}
  \label{app:pairwise_prompt}
\end{figure}

\begin{figure}[!h]
  \centering
  \includegraphics[width=7.8cm, height=13cm]{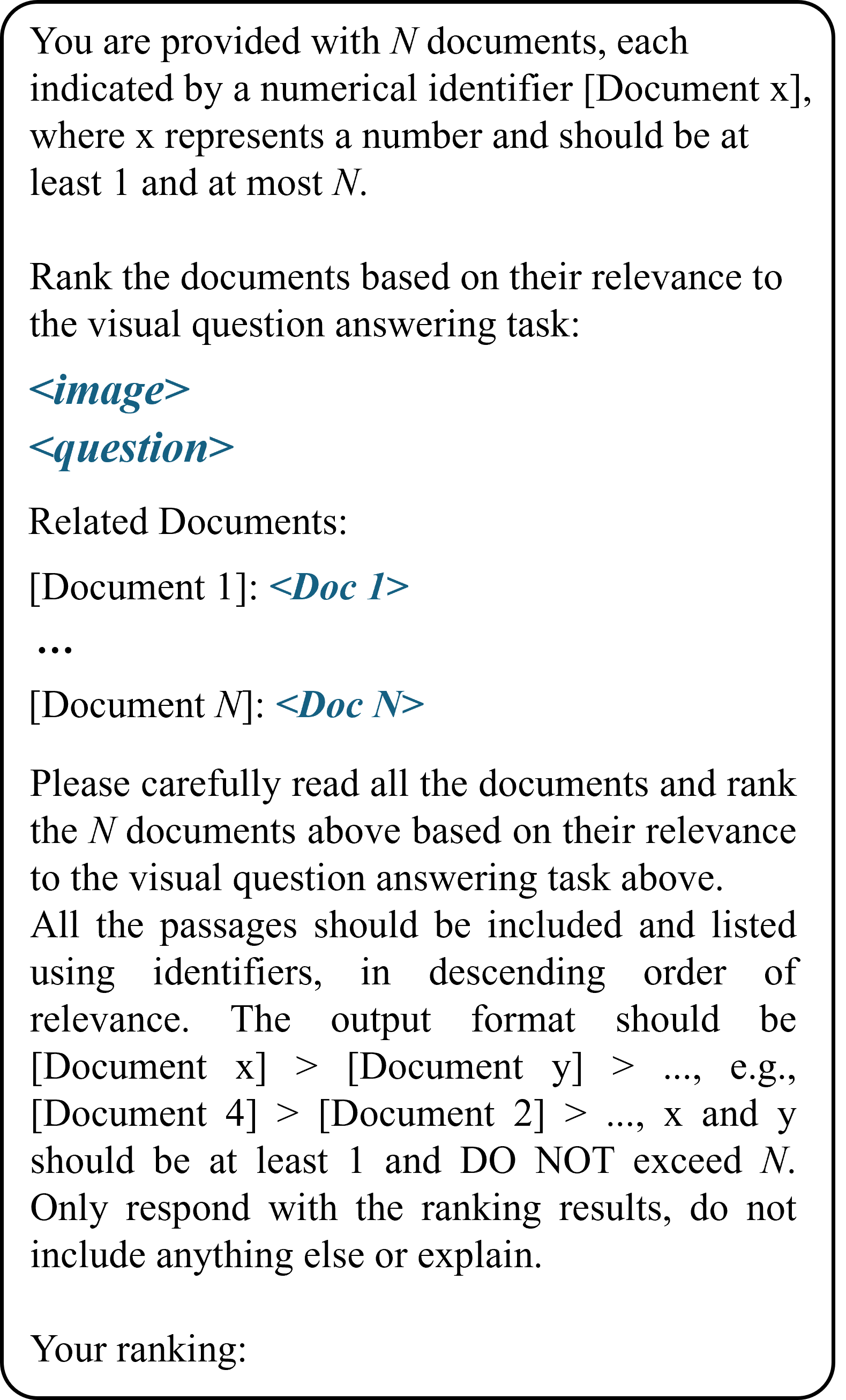}
  \caption{Listwise re-ranking prompt.}
  \label{app:listwise_prompt}
\end{figure}

This section lists the prompt templates used in the re-ranking phase. Figure \ref{app:pairwise_prompt} and \ref{app:listwise_prompt} show the prompt for pairwise and listwise re-ranking, respectively. $N$ is the number of documents to provide.

\subsection{Generation}
\label{generation_prompts}
\begin{figure}[!h]
  \centering
  \includegraphics[width=\linewidth]{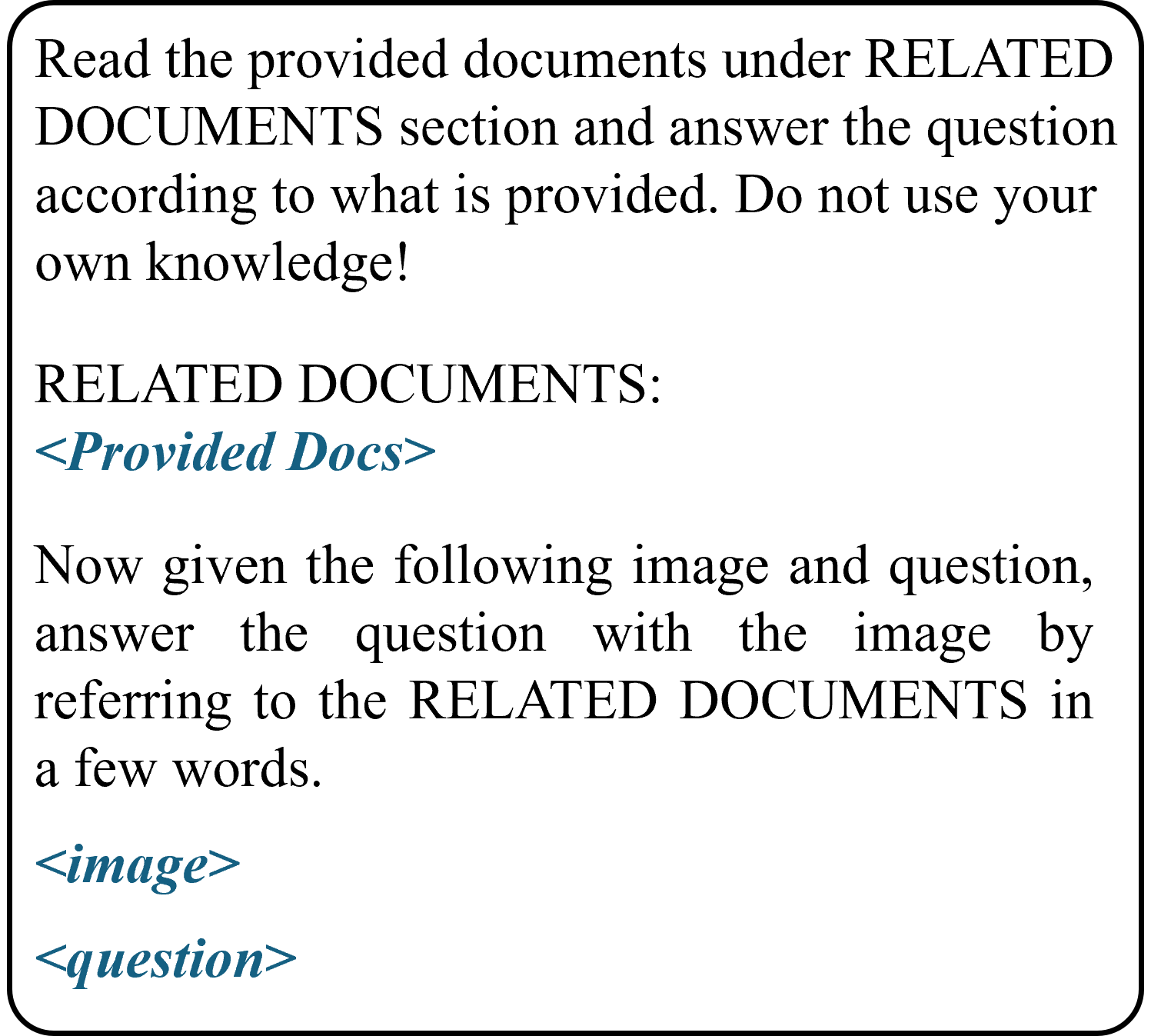}
  \caption{Generation prompt.}
  \label{app:gen_prompt}
\end{figure}

\begin{figure}[!h]
  \centering
  \includegraphics[width=\linewidth]{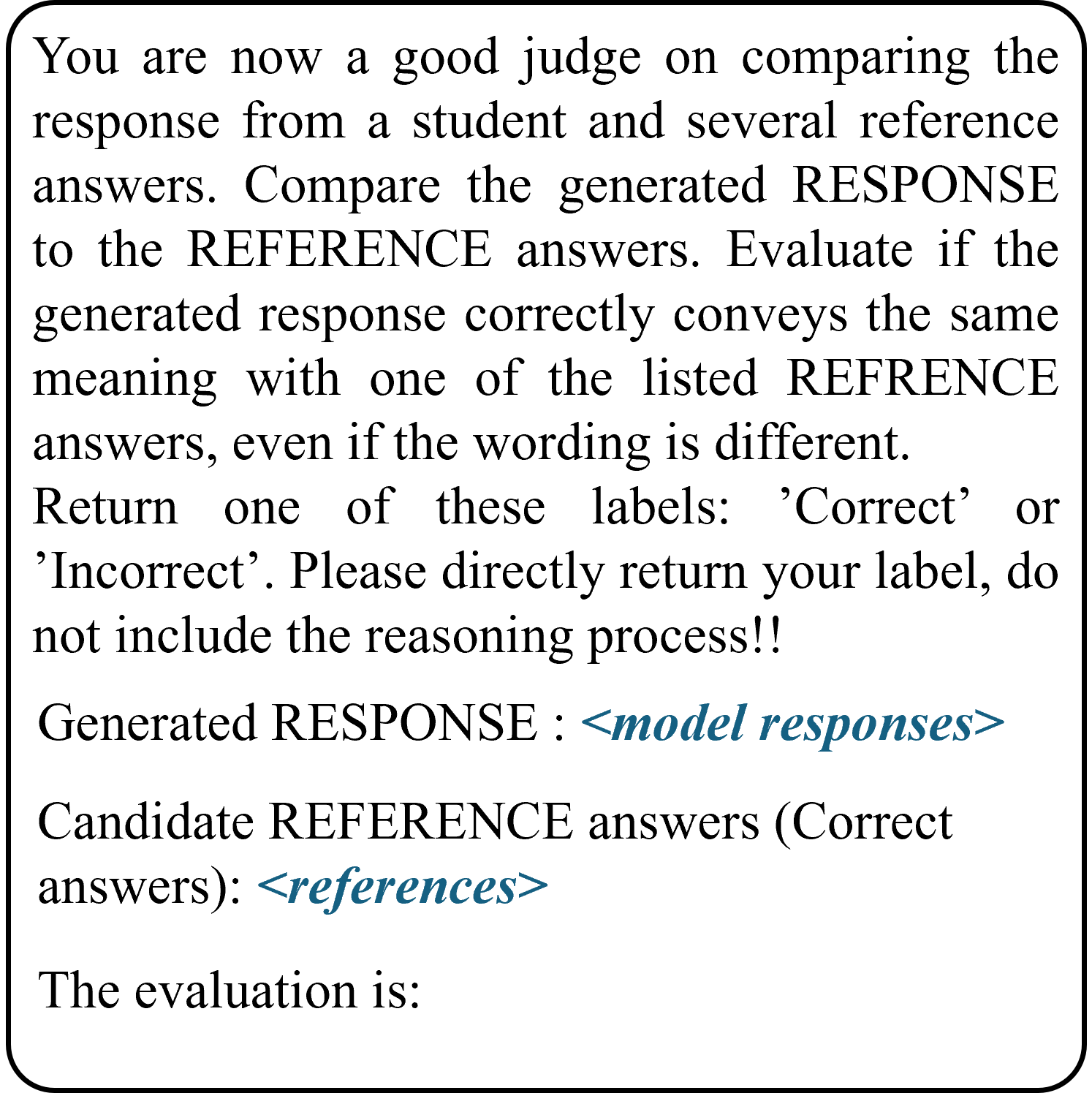}
  \caption{Judge prompt used for the InternVL3 and GPT 4.1.}
  \label{app:judge_prompt}
\end{figure}

This section shows the prompts used during the generation phase. Figure \ref{app:gen_prompt} and \ref{app:judge_prompt} show the prompt for model generation and automated judge, respectively.

\subsection{Unifying re-ranking and generation}
\label{agentic_prompt}
\begin{figure}[!h]
  \centering
  \includegraphics[width=\linewidth]{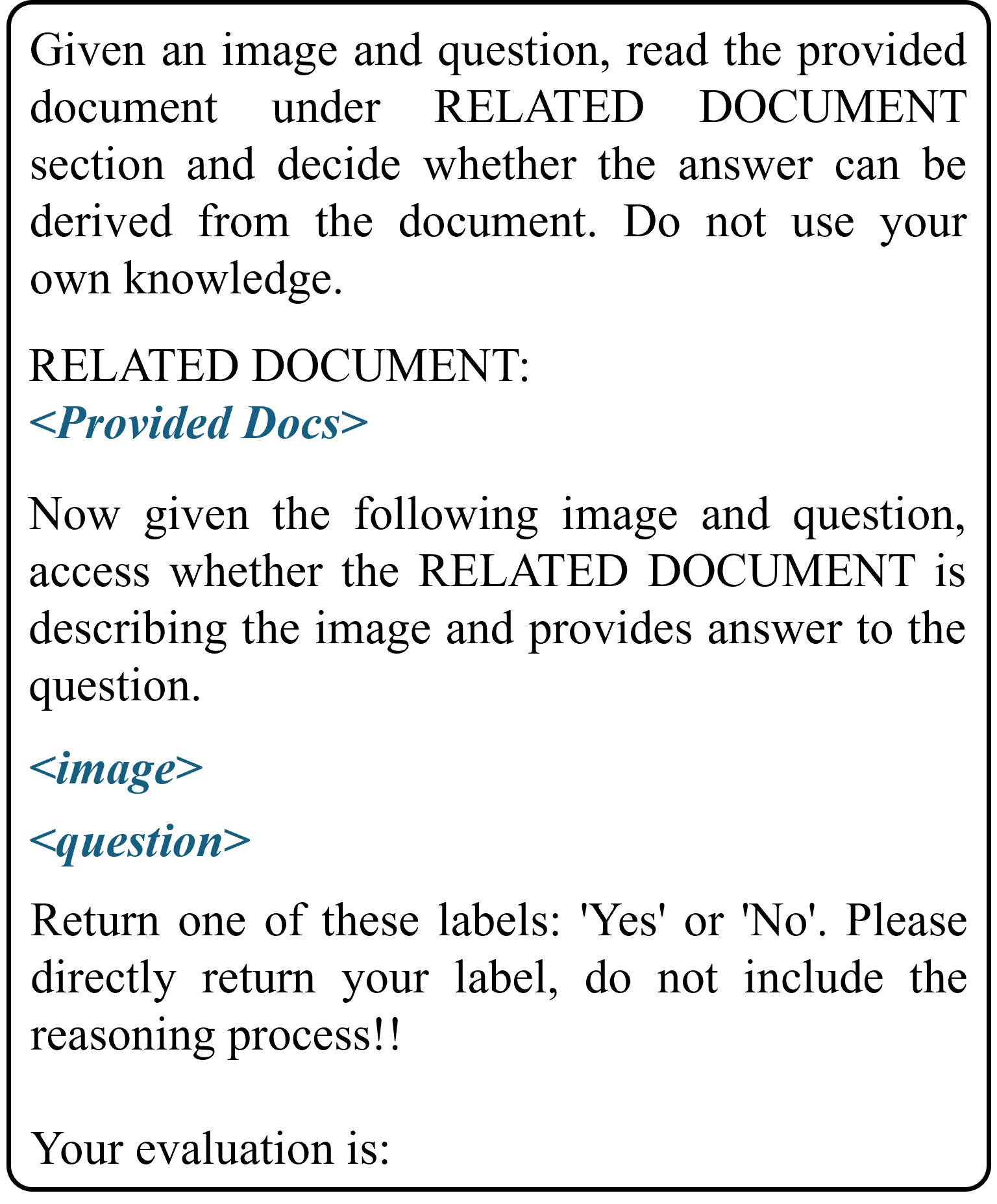}
  \caption{The evaluation prompt to assess the relation between query and provided document \textbf{before the response generation}.}
  \label{app:doc_related_prompt}
\end{figure}

\begin{figure}[!h]
  \centering
  \includegraphics[width=\linewidth]{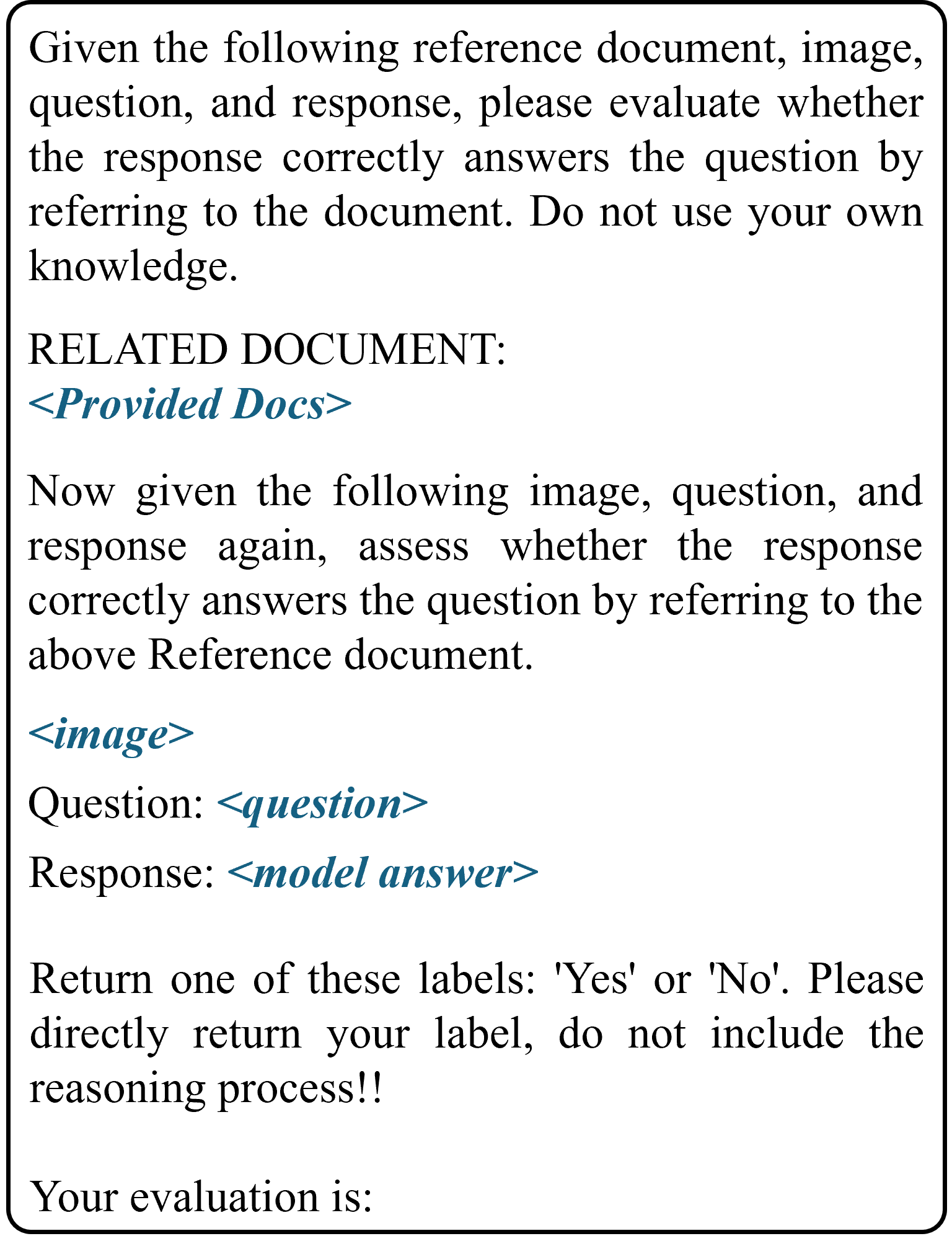}
  \caption{The self-reflection prompt to verify the model response answers to the query and is derived from the provided document.}
  \label{app:self-reflect_prompt}
\end{figure}

Figure \ref{app:doc_related_prompt} outlines the prompt for assessing document relevance to the input query. If the model decides the document is relevant and generates a response, a self-reflection prompt, shown in Figure \ref{app:self-reflect_prompt}, evaluates the validity of the tentative response. Valid response is kept, and an invalid one prompts the model to shift to the following document. If no document is found, the model outputs "Model fails to answer the question", terminating the process.

\section{Experimental Costs}
The cost for the response evaluation using GPT 4.1 cost approximately \$20 in total.


\begin{table*}[t]
\centering
\caption{ROUGE-L score and response accuracy with \textbf{Qwen2-VL-7B-Instruct} as the generation model in Figure \ref{fig:gen_response}.} 
\label{tab:gen_response}
\begin{adjustbox}{width=\textwidth}
\begin{tabular}{c|c|cc|c|cc|c|cc} 
\specialrule{.2em}{.05em}{.05em} 
\textbf{Dataset}                    & \multicolumn{9}{c}{\textbf{Evaluation Methods}}                                                                                                                                                                       \\ 
\specialrule{.1em}{.05em}{.05em}
\specialrule{.1em}{.05em}{.05em} 
\multirow{10}{*}{E-VQA}    & \multicolumn{3}{c|}{ROUGE-L}                                & \multicolumn{3}{c|}{GPT 4.1}                                & \multicolumn{3}{c}{InternVL3}                                                    \\ 
\cline{2-10}
                           & top-k given & \multicolumn{1}{c|}{w/o re-rank} & w/ re-rank & top-k given & \multicolumn{1}{c|}{w/o re-rank} & w/ re-rank & top-k given                     & \multicolumn{1}{c|}{w/o re-rank} & w/ re-rank  \\ 
\cline{2-10}
                           & 1           & 0.3651                           & 0.4063     & 1           & 39.01                            & 41.77      & 1                               & 15.49                            & 17.65       \\
                           & 2           & 0.3786                           & 0.4083     & 2           & 40.11                            & 41.68      & 2                               & 16.4                             & 18.12       \\
                           & 3           & 0.3787                           & 0.3949     & 3           & 40                               & 40.76      & 3                               & 16.55                            & 17.97       \\
                           & 4           & 0.3762                           & 0.3922     & 4           & 39.34                            & 39.71      & 4                               & 16.63                            & 17.91       \\
                           & 5           & 0.3778                           & 0.3919     & 5           & 39.43                            & 39.66      & 5                               & 16.21                            & 17.48       \\ 
\cline{2-10}
                           & lower bound & \multicolumn{2}{c|}{0.1245}                   & lower bound & \multicolumn{2}{c|}{11.05}                    & \multicolumn{1}{c}{lower bound} & \multicolumn{2}{c}{6.23}                       \\
                           & upper bound & \multicolumn{2}{c|}{0.5111}                   & upper bound & \multicolumn{2}{c|}{53.73}                    & \multicolumn{1}{c}{upper bound} & \multicolumn{2}{c}{22.82}                      \\
                           & unified     & \multicolumn{2}{c|}{0.4305}                   & unified     & \multicolumn{2}{c|}{45.66}                    & \multicolumn{1}{c}{unified}     & \multicolumn{2}{c}{19.49}                      \\ 
\specialrule{.1em}{.05em}{.05em} 
\multirow{10}{*}{InfoSeek} & \multicolumn{3}{c|}{ROUGE-L}                                & \multicolumn{3}{c|}{GPT 4.1}                                & \multicolumn{3}{c}{InternVL3}                                                    \\ 
\cline{2-10}
                           & top-k given & \multicolumn{1}{c|}{w/o re-rank} & w/ re-rank    & top-k given & \multicolumn{1}{c|}{w/o re-rank} & w/ re-rank    & top-k given                     & \multicolumn{1}{c|}{w/o re-rank} & w/ re-rank     \\ 
\cline{2-10}
                           & 1           & 0.3905                           & 0.4067     & 1           & 36.92                            & 37.6       & 1                               & 28.38                            & 29.7        \\
                           & 2           & 0.4003                           & 0.4131     & 2           & 37.48                            & 38.2       & 2                               & 29.06                            & 29.91       \\
                           & 3           & 0.3965                           & 0.4101     & 3           & 37.14                            & 37.46      & 3                               & 29.28                            & 30.4        \\
                           & 4           & 0.3978                           & 0.4001     & 4           & 37.68                            & 37.78      & 4                               & 29.12                            & 30.18       \\
                           & 5           & 0.3963                           & 0.408      & 5           & 37.78                            & 38.02      & 5                               & 29.18                            & 30.36       \\
\cline{2-10}
                           & lower bound & \multicolumn{2}{c|}{0.1975}                   & lower bound & \multicolumn{2}{c|}{16.96}                    & lower bound                     & \multicolumn{2}{c}{16.2}                       \\
                           & upper bound & \multicolumn{2}{c|}{0.4659}                   & upper bound & \multicolumn{2}{c|}{44.84}                    & upper bound                     & \multicolumn{2}{c}{33.84}                      \\
                           & unified     & \multicolumn{2}{c|}{0.4151}                   & unified     & \multicolumn{2}{c|}{39.5}                         & unified                         & \multicolumn{2}{c}{30.96}                     \\
\specialrule{.2em}{.05em}{.05em} 
\end{tabular}
\end{adjustbox}
\end{table*}

\begin{table*}[t]
\centering
\caption{ROUGE-L score and response accuracy with \textbf{LLaVA-OneVision} as the generation model in Figure \ref{fig:gen_response}.} 
\label{tab:gen_response_llava}
\begin{adjustbox}{width=\textwidth}
\begin{tabular}{c|c|cc|c|cc|c|cc}
\specialrule{.2em}{.05em}{.05em}
\textbf{Dataset}                    & \multicolumn{9}{c}{\textbf{Evaluation Methods}}             \\
\specialrule{.1em}{.05em}{.05em}
\specialrule{.1em}{.05em}{.05em}
\multirow{10}{*}{E-VQA}    & \multicolumn{3}{c|}{ROUGE-L}                                & \multicolumn{3}{c|}{GPT 4.1}                                & \multicolumn{3}{c}{InternVL3}                                                    \\ 
\cline{2-10}
                           & top-k given & \multicolumn{1}{c|}{w/o re-rank} & w/ re-rank & top-k given & \multicolumn{1}{c|}{w/o re-rank} & w/ re-rank & top-k given                     & \multicolumn{1}{c|}{w/o re-rank} & w/ re-rank  \\ 
\cline{2-10}
                           & 1           & 0.325                            & 0.352      & 1           & 33.56                            & 35.44      & 1                               & 13.6                             & 14.25       \\
                           & 2           & 0.33                             & 0.35       & 2           & 33.52                            & 35.59      & 2                               & 13.87                            & 14.29       \\
                           & 3           & 0.328                            & 0.343      & 3           & 33.62                            & 34.98      & 3                               & 13.7                             & 14.08       \\
                           & 4           & 0.328                            & 0.342      & 4           & 33.53                            & 34.92      & 4                               & 13.95                            & 14.19       \\
                           & 5           & 0.321                            & 0.339      & 5           & 33.87                            & 34.6       & 5                               & 13.68                            & 14.04       \\ 
\cline{2-10}
                           & lower bound & \multicolumn{2}{c|}{0.115}                    & lower bound & \multicolumn{2}{c|}{10.38}                    & \multicolumn{1}{c}{lower bound} & \multicolumn{2}{c}{6.19}                       \\
                           & upper bound & \multicolumn{2}{c|}{0.409}                    & upper bound & \multicolumn{2}{c|}{41.54}                    & \multicolumn{1}{c}{upper bound} & \multicolumn{2}{c}{17.79}                      \\
                           & unified     & \multicolumn{2}{c|}{0.372}                    & unified     & \multicolumn{2}{c|}{40.69}                    & \multicolumn{1}{c}{unified}     & \multicolumn{2}{c}{16.57}                      \\ 
\specialrule{.1em}{.05em}{.05em}
\multirow{10}{*}{InfoSeek} & \multicolumn{3}{c|}{ROUGE-L}                                & \multicolumn{3}{c|}{GPT 4.1}                                & \multicolumn{3}{c}{InternVL3}                                                    \\ 
\cline{2-10}
                           & top-k given & \multicolumn{1}{c|}{w/o re-rank} & w/ re-rank & top-k given & \multicolumn{1}{c|}{w/o re-rank} & w/ re-rank & top-k given                     & \multicolumn{1}{c|}{w/o re-rank} & w/ re-rank  \\ 
\cline{2-10}
                           & 1           & 0.361                            & 0.376      & 1           & 34.66                            & 35.28      & 1                               & 27.4                             & 28.68       \\
                           & 2           & 0.362                            & 0.372      & 2           & 35.02                            & 36.11      & 2                               & 27.9                             & 28.76       \\
                           & 3           & 0.352                            & 0.367      & 3           & 34.77                            & 35.66      & 3                               & 26.9                             & 28.18       \\
                           & 4           & 0.352                            & 0.364      & 4           & 33.14                            & 34.5       & 4                               & 26.76                            & 28.1        \\
                           & 5           & 0.349                            & 0.36       & 5           & 33.38                            & 34.12      & 5                               & 26.72                            & 27.8        \\ 
\cline{2-10}
                           & lower bound & \multicolumn{2}{c|}{0.143}                    & lower bound & \multicolumn{2}{c|}{12.12}                         & \multicolumn{1}{c}{lower bound} & \multicolumn{2}{c}{11.18}                      \\
                           & upper bound & \multicolumn{2}{c|}{0.427}                    & upper bound & \multicolumn{2}{c|}{40.4}                         & \multicolumn{1}{c}{upper bound} & \multicolumn{2}{c}{32.24}                      \\
                           & unified     & \multicolumn{2}{c|}{0.385}                    & unified     & \multicolumn{2}{c|}{37.86}                    & \multicolumn{1}{c}{unified}     & \multicolumn{2}{c}{29.72}                      \\
\specialrule{.2em}{.05em}{.05em} 
\end{tabular}
\end{adjustbox}
\end{table*}
\section{Licenses}
The datasets we used, InfoSeek and E-VQA, are licensed under Apache License 2.0 and CC BY 4.0, respectively.
The retrieval models, CLIP, EVA-CLIP, BGE-CLIP, BLIP, BGE-MLLM, and GME, are under MIT License, MIT License, MIT License, MIT License, MIT License, Apache license 2.0, correspondingly.
The re-ranker models, MM-Embed \cite{mm-embed} and Qwen2-VL-7B-Instruct, are licensed under CC-BY-NC-4.0 and Apache License 2.0. Q-former from EchoSight \cite{echosight} was released without an accompanying license.
The response generation models, Qwen2-VL-7B-Instruct and LLaVA-OneVision, are both licensed under Apache License 2.0.
The InternVL3-14B model is released under MIT License. 

Our use of the released datasets and models are all consistent with their intended use.

\end{document}